\documentclass[sigconf]{acmart}
\AtBeginDocument{%
  \providecommand\BibTeX{{%
    \normalfont B\kern-0.5em{\scshape i\kern-0.25em b}\kern-0.8em\TeX}}}

\copyrightyear{2023}
\acmYear{2023}
\setcopyright{acmlicensed}
\acmConference[KDD '23] {Proceedings of the 29th ACM SIGKDD Conference on Knowledge Discovery and Data Mining}{August 6--10, 2023}{Long Beach, CA, USA.}
\acmBooktitle{Proceedings of the 29th ACM SIGKDD Conference on Knowledge Discovery and Data Mining (KDD '23), August 6--10, 2023, Long Beach, CA, USA}
\acmPrice{15.00}
\acmISBN{979-8-4007-0103-0/23/08}
\acmDOI{10.1145/3580305.3599760}
\usepackage{subfigure,multirow,enumitem} 
\usepackage[ruled,vlined]{algorithm2e}
\begin{document}
\title{A Data-driven Region Generation Framework for Spatiotemporal Transportation Service Management}


\author{Liyue Chen}
\affiliation{%
  \institution{Key Lab of High Confidence Software Technologies (Peking University), Ministry of Education \& \\
  School of Computer Science, \\ Peking University}
  \state{Beijing}
  \country{China}
}
\email{chenliyue2019@gmail.com}

\author{Jiangyi Fang}
\affiliation{%
  \institution{School of Artificial Intelligence and Automation, Huazhong University of Science and Technology}
  \city{Wuhan}
  \country{China}}
\email{fangjiangyi2001@gmail.com}

\author{Zhe Yu}
\affiliation{%
  \institution{DiDi Chuxing}
  \city{Hangzhou}
  \country{China}}
\email{eugeneyu@didiglobal.com}

\author{Yongxin Tong}
\affiliation{%
   \institution{State Key Laboratory of Software Development Environment, Beijing Advanced Innovation Center for Future Blockchain and Privacy Computing, Beihang University}
\city{Beijing}
\country{China}
}
\email{yxtong@buaa.edu.cn}

\author{Shaosheng Cao}
\affiliation{%
  \institution{DiDi Chuxing}
  \city{Hangzhou}
  \country{China}}
\email{shelsoncao@didiglobal.com}
\authornotemark[1]

\author{Leye Wang}
\affiliation{%
  \institution{Key Lab of High Confidence Software Technologies (Peking University), Ministry of Education \& \\
  School of Computer Science, \\ Peking University}
  \state{Beijing}
  \country{China}
}
\email{leyewang@pku.edu.cn}
\authornote{Corresponding authors.}

\renewcommand{\shortauthors}{Chen et al.}


\begin{abstract}
MAUP (modifiable areal unit problem) is a fundamental problem for spatial data management and analysis. As an instantiation of MAUP in online transportation platforms, region generation (i.e., specifying the areal unit for service operations) is the first and vital step for supporting spatiotemporal transportation services such as ride-sharing and freight transport. Most existing region generation methods are manually specified (e.g., fixed-size grids), suffering from poor spatial semantic meaning and inflexibility to meet service operation requirements. In this paper, we propose \textbf{\textit{RegionGen}}, a data-driven region generation framework that can specify regions with key characteristics (e.g., good spatial semantic meaning and predictability) by \textit{modeling region generation as a multi-objective optimization problem}. First, to obtain good spatial semantic meaning, \textit{RegionGen} segments the whole city into atomic spatial elements based on road networks and obstacles (e.g., rivers). Then, it clusters the atomic spatial elements into regions by maximizing various operation characteristics, which is formulated as a multi-objective optimization problem. For this optimization problem, we propose a multi-objective co-optimization algorithm. Extensive experiments verify that \textit{RegionGen} can generate more suitable regions than traditional methods for spatiotemporal service management.
\end{abstract}

\ccsdesc[500]{Information systems~Spatial-temporal systems}
\ccsdesc[500]{Applied computing~Transportation}

\keywords{Modifiable areal unit; spatial data management}



\maketitle

\section{Introduction} \label{introduction}
The global transportation services market is expected to grow to over 7,200 billion and 10 trillion dollars in 2022 and 2026 over a compound annual growth rate of 9\% according to the market report \cite{transport_market_report}. For online transportation platforms (e.g., Uber), 
\textit{region is the fundamental operation areal unit for spatiotemporal data management}. 
For example, aiming to shorten passengers' waiting time, online transportation platforms first divide the whole city into several regions (e.g., $1km\times 1km$ grids \cite{zhang2017deep,geng2019spatiotemporal}) and dispatch idle drivers to hot areas based on the estimated demand for each region.
Only with a well-managed set of regions, online transportation platforms can dispatch more orders to drivers and respond to passengers' ride-sharing requests more quickly.

\begin{figure*}[t]
  \centering
  \includegraphics[width=.9\linewidth]{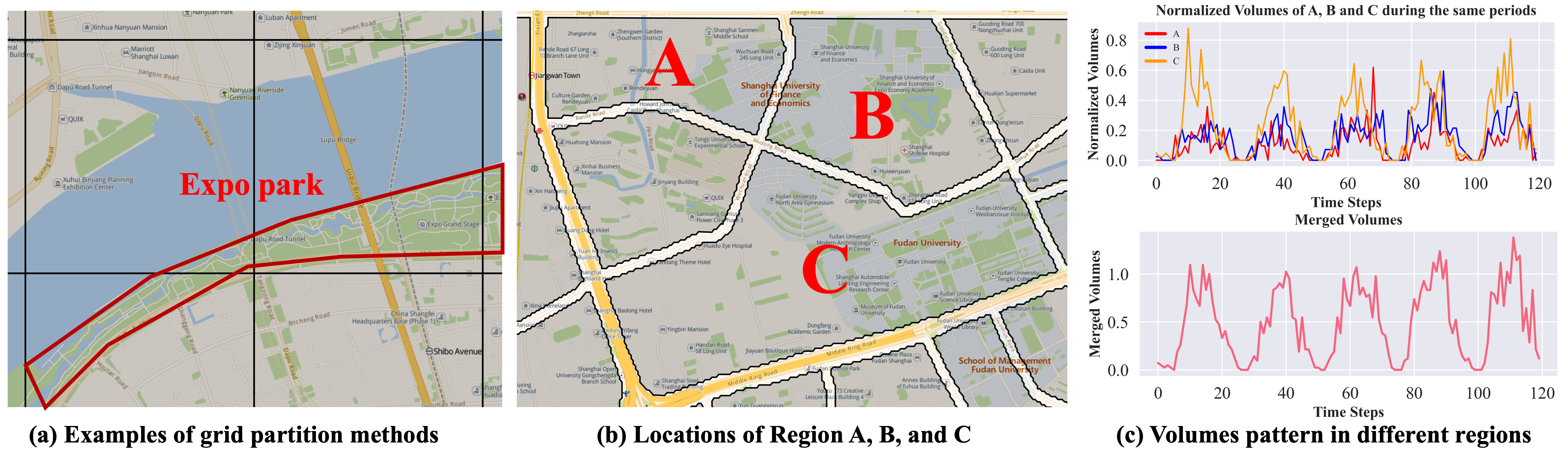}
  \vspace{-.6em}
  \caption{(a): Examples of grid partition methods that generate regions with poor spatial semantic meaning. The main green part is a park. (b) and (c): Generated by road segmentation methods, Region A, B, and C are with poor predictability, which may be improved by merging these three regions.}
  \label{fig:pitfall}
  \vspace{-0.6em}
\end{figure*}

In general, how to determine the region (i.e., areal unit) for spatial data management is one of the most important problems in geo-science, widely known as the \textit{the modifiable areal unit problem} (MAUP) \cite{wong2004modifiable,openshaw1981modifiable}, since different region specifications may lead to diverged analysis outcomes \cite{de2021multicriteria}. 
Despite the importance of the region specification, current industry practice is mostly ad-hoc and manually defined, e.g., \textit{fixed shapes like grids or hexagons} \cite{zhang2017deep,ke_short-term_2017,geng2019spatiotemporal,sthan_2022}, and \textit{street/administrative blocks} \cite{zheng_taxicabs,liu_causal,yuan_functions_2012,city_noise_2014,zhang_multi_view_2020}, without comprehensive assessment standards.
This could incur uncontrollable and unanticipated effects on the operations of spatiotemporal services. As later shown in our experiments, compared to ad-hoc grid regions, an optimized region specification can reduce the spatiotemporal service demand prediction error by around $10\%$; this improvement is particularly noteworthy considering that a recent benchmark study \cite{STMeta} has shown that state-of-the-art deep models can improve prediction performance also by about $10\%$ compared to classical methods (e.g., gradient boosted regression trees \cite{li_traffic_2015}).
Hence, region specification is a crucial but largely under-investigated issue for spatiotemporal data management practice.
Next, we briefly illustrate existing region specification methods and their pitfalls.

Typically, \textit{fixed-shape} methods divide the whole city into numerous regions with the same shape, such as grids and hexagons, \textbf{in the absence of geographic semantic meaning}.\footnote{Regions with good semantic meaning are bounded by roads \cite{yuan2012segmentation}.} As demonstrated in Fig. \ref{fig:pitfall}(a), the Shanghai Expo Park is an entire geographic entity, but it is separated into several grids, violating the semantics of urban spaces. In addition, crossing a river usually takes longer for both automobiles and pedestrians than crossing a road. However, the fact that the two sides of the river are divided into the same grid makes it inconvenient to operate online transportation services. For instance, when a driver is guided to such a grid to wait for ride-sharing requests, she/he would be doubtful whether she/he needs to go across the river or not.

\textit{Street/administrative-block} methods use pre-defined spatial boundaries (i.e., roads and administrative boundary) for region generation.
While these methods can keep good geographic semantic meaning compared to \textit{fixed-shape} methods, they generally \textbf{lack the flexibility to well fit various operation requirements} of online transportation services.
For instance, administrative blocks may be too coarse to enable fine-grained operations (especially for downtown areas with a large number of demands); meanwhile, road-segmented street blocks may be too small and thus most blocks hold nearly zero demands.

A practically key spatiotemporal data management question then emerges: \textbf{can we generate a good set of regions for online transportation platforms by meeting two goals: (i) with good geographic semantic meaning, and (ii) flexibly fitting various operation requirements of spatiotemporal services?}

To address the above problem, we propose and implement a unified region generation framework, called \textit{\textbf{RegionGen}}, which enables the adaptive formation of regions for various spatiotemporal transportation services, such as \textit{taxi dispatching}, \textit{freight transportation}, and \textit{designated driver} services. 
The general process of \textit{RegionGen} follows two steps. In the first step, we adapt road segmentation techniques \citep{yuan2012segmentation} to extract semantically-meaningful atomic spatial units.
While the atomic spatial units may be too small for operation, the second step merges nearby units into larger regions, which can be seen as a spatial clustering process. 

In particular, our spatial clustering process needs to consider various realistic operation characteristics for spatiotemporal services, including generated regions' granularity, specificity, etc. \cite{li_traffic_2015,chen_dynamic_2016,rebalancing_2019,hu_partition_2019}. 
More specifically, we argue that the \textit{temporal predictability} plays a key role in transportation service operations.
If the regions are generated with higher predictability, future events of interests (e.g., the number of ride-sharing requests in next one hour) are easier to forecast and thus service operation decisions can be made more accurately (e.g., how many drivers need to be dispatched to a certain region).
Fig. \ref{fig:pitfall}(b) and (c) illustrate a toy example to show how spatial clustering can help generate a high-predictability region for freight services. 
A, B, and C are three atomic spatial units generated by road segmentation; nevertheless, their temporal patterns on freight are unstable, as each spatial unit contains fewer data and are susceptible to random disturbance.
Then, performing operation management directly based on these three spatial units will be difficult.
Meanwhile, we find that these three spatial units all contain university campuses;
then, by clustering them together for operation, we obtain a highly-predictable merged region whose temporal regularity is significantly increased. 

In summary, our main contributions include:
\begin{itemize}
\item To the best of our knowledge, this is one of the pioneering efforts toward generating regions considering key operation characteristics (e.g., good spatial semantic meaning and predictability) in spatiotemporal service. This is also one of the first data-driven solutions to MAUP for spatiotemporal transportation service operations. 

\item \textit{RegionGen} includes two main steps. First, to keep good spatial semantic meanings, \textit{RegionGen} segments the whole city into atomic spatial elements based on road networks and obstacles (e.g., rivers). Second, it clusters the atomic spatial elements by maximizing the key operating characteristics with a multi-objective optimization process. 
	
\item We conduct extensive experiments on three spatiotemporal service datasets. Results verify that \textit{RegionGen} can generate more suitable regions than traditional methods for spatiotemporal transportation service.
\end{itemize}

\section{Operation Characteristics Analysis and Problem Formulation}
There are two main challenges in designing such a region generation framework. The first is to guarantee the generated regions with good spatial semantic meaning. 
The second is to make the generated regions meet the operation requirements of spatiotemporal services. Our basic idea is to aggregate segmented fine-grained regions into large ones to acquire better operating characteristics (e.g., predictability). It is worth noting that the aggregated regions are still bounded by roads and not cross obstacle entities, and thus still guarantee good spatial semantics. 

Accordingly, we disassemble the region generation process in two steps. The first step is to generate many fine-grained regions with good spatial semantics by segmentation (in the rest of the paper, for clarity, the fine-grained regions are called \textbf{atomic spatial elements}). 
Then, the second step is to aggregate atomic spatial elements to obtain better operating characteristics. Especially, we first quantify several spatiotemporal services operating characteristics (Sec. \ref{operating_characteristics}). While there exist a variety of spatiotemporal services in practice, we have summarized three types of general operating characteristics (i.e., predictability, granularity, and specificity) that may benefit most services.

\subsection{Quantifying Operation Characteristics} \label{operating_characteristics}
In most urban prediction applications (e.g., ride-sharing or dockless bike-sharing demand prediction), we usually expect regions with pleasant \textbf{predictability}, fine-grained spatial \textbf{granularity}, and high service \textbf{specificity}. \textit{Predictability} refers to whether future events of interest for the service (e.g., ride-sharing demands) can be easily predicted. 
At the same time, many spatiotemporal services (e.g., traffic monitoring) hope that the spatial \textit{granularity} can be fine-grained (e.g., the region size is not too large) to carry out precise management. Besides, high \textit{specificity} indicates that the generated region closely matches the actual service area, i.e., the generated region has few redundant parts where (almost) no service is required (e.g., ride-sharing service operation regions may not need to cover mountain areas where cars cannot reach). We here quantify these general characteristics for the region generation process.

\textbf{Predictability.} As there exist various prediction models that can be adopted in practice \cite{STMeta}, it is generally non-trivial to efficiently and precisely quantify the predictability of the spatiotemporal time-series data within regions.
An intuitive way could be firstly fixing a prediction model and then measuring the prediction errors on test records (e.g., mean absolute error, Kaboudan metric \cite{Kaboudan_1999}).
Such measurements are highly model-dependent. 
That is, the error measurements obtained by one model cannot usually represent another model.
As state-of-the-art spatiotemporal prediction models are mostly complex deep models \cite{STMeta}, training these models to measure predictability for every possible region generation candidate would be too time-consuming and thus practically intractable.

We then propose to use \textit{model-agnostic} measures to efficiently estimate regions' predictability without the need to repeatedly training deep models.
Specifically, we select the auto-correlation function (ACF) as a fast proxy measurement for predictability. Suppose that $\{s_{1,i},s_{2,i},...,s_{T,i}\}$ is the time series of region $i$, the ACF of region $i$ after $k$ slots delay is computed by:
\begin{equation}
\rho^k_i=\frac{T\cdot\sum_{{t=k+1}}^T(s_{t,i}-\bar{s_i})(s_{t-k,i}-\bar{s_i})}{(T-k)\cdot \sum_{t=1}^T(s_{t,i}-\bar{s_i})^2}
\label{eq:acf}
\end{equation}
where $\bar{s_i}$ is the mean value of the time series in region $i$. ACF is often used to measure the periodicity of time series data. While high periodicity (e.g., daily periodicity) is one of the dominant indicators for accurate prediction \cite{STMeta}, we deem that high periodicity may be highly correlated with low prediction errors of deep models. To investigate whether such correlations exist, we explore the relationship between the ACF after 24 hours delay (called ACF\_daily) and prediction errors regarding a state-of-the-art deep forecasting model (i.e., \textit{STMGCN} \cite{geng2019spatiotemporal}) in Fig. \ref{fig:acf_predictability} (Appendix~\ref{app:acf}). 
The results verify that ACF\_daily and prediction errors are highly correlated, especially when ACF\_daily is large; that is, maximizing the ACF\_daily would practically decrease the prediction errors significantly. Therefore, in this work, we adopt the ACF\_daily indicator for efficiently measuring the predictability of regions.

\textbf{Granularity.} In real-world applications, to keep good spatial semantics, it is difficult to enforce the generated region with specific shapes and we typically measure the region size by the region area. More importantly, keep in mind that the greater the area, the more spatiotemporal data it contains, and the less random noise in the time series, the more regular the time series are, and the more predictable they are. Therefore, unlike predictability, we cannot require that the generated regions be as small as possible. As a result, we need to make a trade-off between good predictability and fine-grained spatial granularity. In practice, we typically meet the need for fine-grained spatial granularity by imposing the maximum region area constraint. Suppose that $P_i$ is the geographic shape of region $i$, stored as a set of vector coordinates, the granularity of region $i$ is quantified by its area:
\begin{equation}
    ts_i = \text{Area}(P_i)
\end{equation}
where $\text{Area}(\cdot)$ is the area of the 2D polygon calculation function, having been widely integrated into tools, e.g., ArcGIS.\footnote{https://www.arcgis.com/}

\textbf{Specificity.} 
For most spatiotemporal services, the regions for operation management do not always need to cover the entire target area (e.g., a city), as many sub-areas may not have the service requirements (e.g., ride-sharing services may not be required for mountain sub-areas where vehicles cannot access).
To this end, the generated regions suitable for operation management would prefer to cover only those service-required sub-areas. We thus propose the \textit{specificity} to measure the ratio of service-required sub-areas within the generated regions.
Then, low-specificity regions mean that there exist a lot of redundant sub-areas that do not have service requirements and may negatively impact the operation efficiency. For example, when providing online ride-sharing services, if the service specificity of a region is low, the ride-sharing demands are mainly concentrated in only a few hotspots. When dispatching idle drivers to this region, a large number of drivers may influx to the same place, causing traffic congestion. 

Particularly, we define the ratio of the service-required area ($vs$) to the total area of the region ($ts$) as the \textit{specificity}, where the service-required area is counted according to historical service records. While spatiotemporal data records usually store spatial information in the format of points (e.g., latitude and longitude), we first convert these data points to Geohash \cite{geohashwiki} units for further area size calculation. Especially, we use 8-bit geohash (Fig. \ref{fig: geohash_details} in Appendix \ref{app:example_spec} displays the geohash in the service-required area and the whole area) as the calculation unit to get an efficient and effective approximation to the area size. The service specificity of the region $i$ is calculated as:
\begin{equation}
    c_i = \frac{vs_i}{ts_i} \approx \frac{\text{\# of historically-serviced geohash in region } i}{\text{\# of total geohash in region } i}
\end{equation}

\subsection{Problem Formulation} \label{clustering_problem}
\subsubsection{Atomic Spatial Element Segmentation Problem}
Given road networks and geographic obstacles (e.g., rivers), the segmentation problem aims to generate road segments bounded by roads and not overlapping with obstacles.

\subsubsection{Atomic Spatial Element Clustering Problem}
\label{sub:problem_clustering}

Given a graph $G$ of $N$ nodes (i.e., the atomic spatial elements $\mathcal{P}$), the adjacency matrix $A\in  \mathbb{R}^{N\times N}$ (details in~\ref{adjacent_matrix}), spatiotemporal raster data $D\in \mathbb{R}^{T\times N}$ which is extracted from $\mathcal{P}$ and historical service records $\mathcal{D}=\{(l_i,t_i)\}$ ($l_i$ and $t_i$ are the locations and the timestamp that the service takes place), the number of time intervals $T$, maximum area constraint $L$, and the service-required area and total area $vs, ts \in \mathbb{R}^{N\times 1}$ of atomic spatial elements, we aim to cluster $N$ atomic spatial elements to $M$ clusters $(2 \leq M < N)$ by maximizing the average predictability and service specificity of clusters.
\begin{align}
\text{Maximize}\ & f_1(X)= \frac{1}{M}\sum_{j=1}^M\rho_j^{daily}  \label{obj:acf} \\
\text{Maximize}\ & f_2(X)= \frac{1}{M}\sum_{j=1}^M\frac{\sum_{i=1}^Nx_{i,j}\cdot vs_i}{\sum_{i=1}^Nx_{i,j}\cdot ts_i} \label{obj:specificity}\\
\text{s.t.}\ & \sum_{j=1}^Mx_{i,j} =1  \ \ \ \forall i \in [N] \label{element_constraint}\\
& \sum_{i=1}^Nx_{i,j} \ge 1 \ \ \ \forall j \in [M] \label{cluster_constraints} \\ 
& \sum_{i=1}^Nx_{i,j}\cdot ts_i \le L  \ \ \ \forall j \in [M]\label{area_constraints} \\
& x_{u,i}+x_{v,i} - \sum_{z\in \mathcal{S}}x_{z,i} \leq 1 \ \ \ \forall A_{u,v}=0, \mathcal{S}\in \Gamma(u,v) \label{connected_constraints}
\end{align}
where $X\in \{0,1\}^{N\times M}$ is the binary clustering results. $x_{i,j}=1$ denotes $i^{th}$ atomic spatial element belong to $j^{th}$ cluster. $S=D \times X \in \mathbb{R}^{T \times M}$ is the aggregated ST raster data. $\rho_j^{daily}$ is the ACF of the time series in the $j^{th}$ cluster after one day delay. Eq. \ref{element_constraint} impose each atomic spatial element can only be assigned to at most one cluster. Ineq. \ref{cluster_constraints} ensures that each cluster contains at least one atomic spatial element. The overall area of each aggregated cluster must be less than the specified maximum area (Ineq. \ref{area_constraints}). Ineq. \ref{connected_constraints} guarantees that every cluster induces a connected subgraph, where $u,v$ are two non-adjacent nodes in $G$. A set $\mathcal{S} \subseteq V\backslash \{u,v\}$ is a $(u,v)$-separator if $u$ and $v$ belong to different components of $G-S$. We denote by $\Gamma(u,v)$ the collection of all minimal $(u,v)$-separators in $G$ \cite{MIYAZAWA2021826}.

\textbf{Remark}. Eq.~\ref{obj:acf} - Eq.~\ref{connected_constraints} present a bi-objective optimization problem to maximize both the specificity and predictability of clustered regions. The optimization problem is NP-hard as it could be reduced to the well-known balanced graph partition problem, considering a special case that the optimized clustering result may be a balanced partition based on the data volume.\footnote{In practice, it is possible as more data often leads to better predictability (Fig.~\ref{fig: data_acf} in Appendix) and balanced partitioned clusters may have high predictability on average.} Hence, approximation algorithms or heuristics are needed to be designed to find a good solution in a reasonable amount of time.

\section{Method}
\subsection{Framework Overview}
\begin{figure}[htbp]
\centering
\includegraphics[width=1\linewidth]{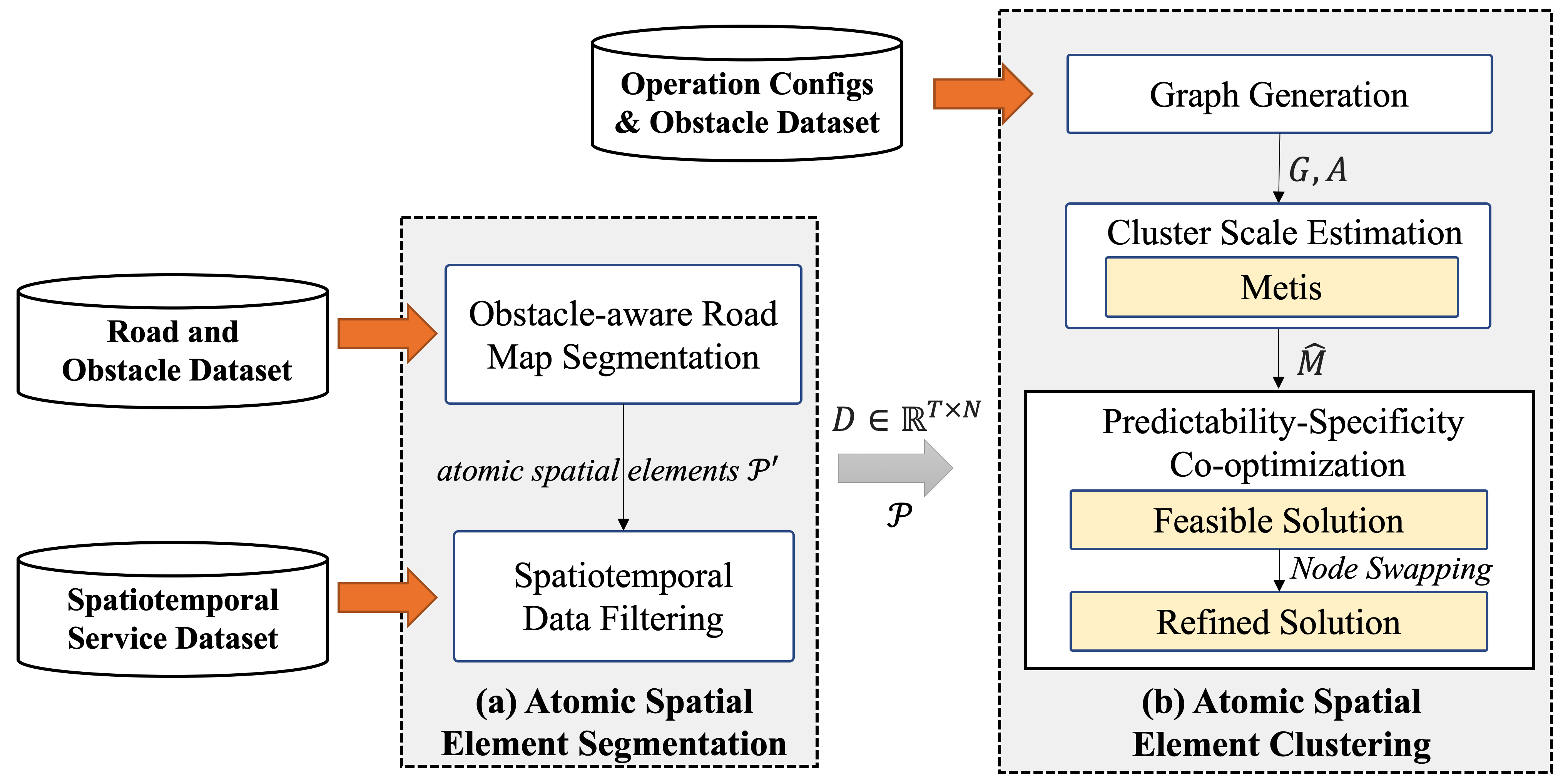}
\caption{Overview of \textit{RegionGen} framework.}
\label{fig: framework}
\end{figure}

The proposed region generation framework \textit{RegionGen} consists of two core components, including the atomic spatial element segmentation and clustering (Fig. \ref{fig: framework}). The atomic spatial elements segmentation component first extracts atomic spatial elements with good spatial semantic meaning by using the proposed obstacle-aware road map segmentation techniques. Then, the spatiotemporal data filtering block removes the atomic spatial elements with less service data to reduce the scale of the subsequent clustering problem.
The atomic spatial elements clustering component first represents the atomic element clustering problem in graph formats, where the nodes are the atomic elements. By combining domain knowledge, we establish the edge sets, which indicate that related nodes can be aggregated into the same cluster. The cluster scale estimation component provides a minimal cluster number that enables the aggregated clusters to meet the operation requirements (i.e., granularity in Eq.~\ref{area_constraints}), allowing for adapting the well-researched graph partition approaches to address the clustering problem that requires the fixed partition number (i.e., clustering number) as inputs. Finally, the predictability-specificity co-optimization component generates regions by solving the atomic spatial element clustering problem. 

\subsection{Atomic Spatial Element Segmentation} \label{road_map_segmentation}
\subsubsection{Obstacle-aware Road Map Segmentation}
Road segments provide us with more natural and semantic meaning than \textit{fixed-shape} methods. In the real world, geographic entities (e.g., parks and residential areas) are bounded by roads and people live in these roads-segmented regions and POIs (Points of Interests) fall in these regions instead of the main roads \cite{yuan2012segmentation}. Previous research \cite{liu_causal,zheng_taxicabs, yuan2012segmentation,yuan_functions_2012} adopt image-based road segmentation techniques that mainly consist of three morphological operators, namely dilation, thinning, and connected component labeling (CCL). Dilation is designed to remove some redundant road details for segmentation avoiding the small connected areas induced by these unnecessary details (e.g., the lanes of roads and the overpasses). The thinning operator aims to extract the skeleton of the dilated road segments while keeping the topology structure of the original image. The CCL operator finds the connected pixels with the same label in the image and eventually generates road segmentation. 

\begin{figure}[htbp]
\centering
\includegraphics[width=.67\linewidth]{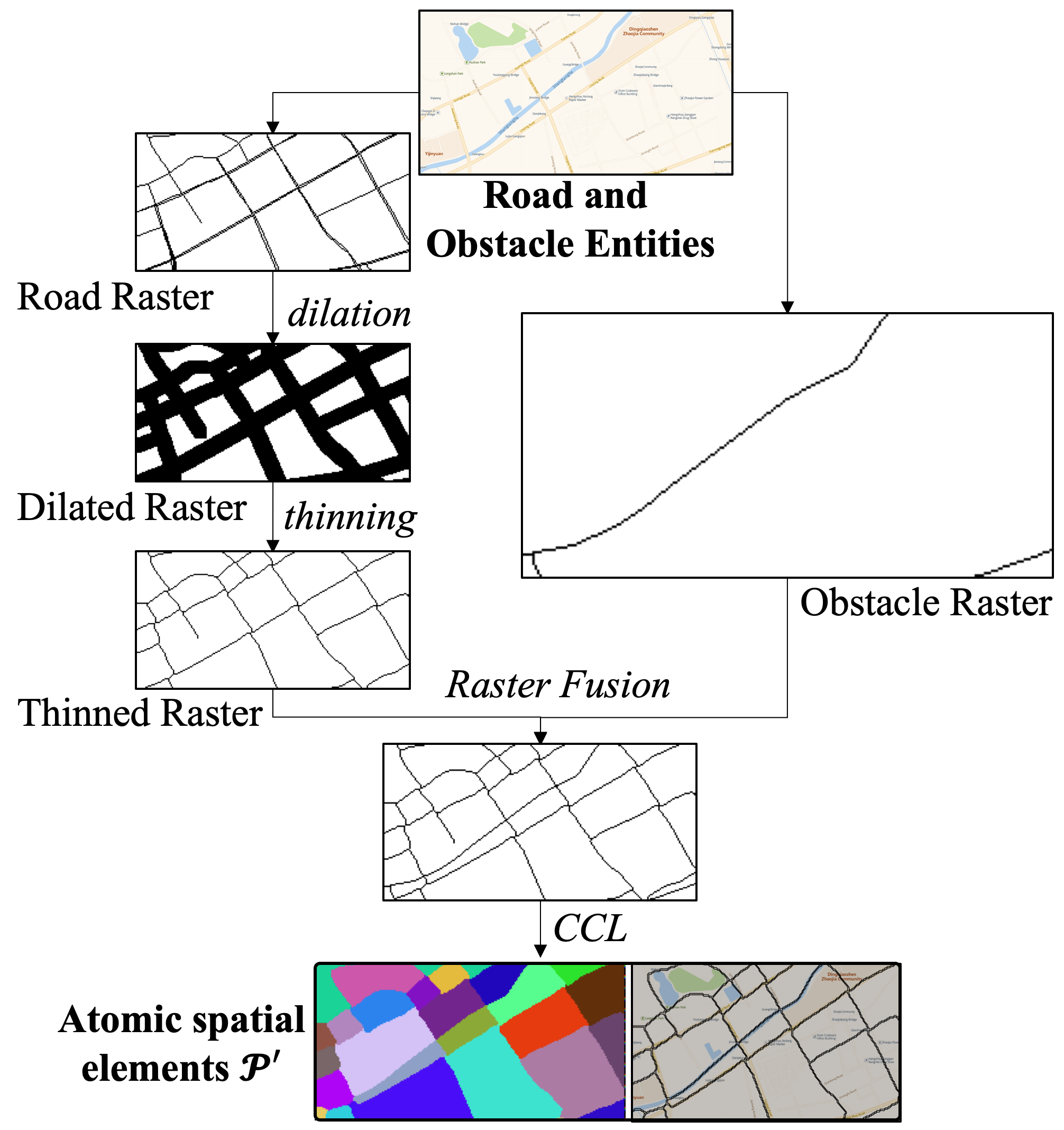}
\caption{Obstacle-aware road map segmentation.}
\label{fig:procedure_seg}
\end{figure}

Although these road segmentation techniques can generate regions bounded with roads and keep good spatial semantic meaning. We argue that the generated regions may still be inconvenient for operating spatiotemporal services since the regions may cross obstacle entities (e.g., rivers). To overcome this shortage, we proposed the obstacle-aware road map segmentation method that incorporates obstacle entities into the road map segmentation process, enabling the road segmentation will not to stretch over obstacle entitles and thus be more suitable for operating services.

Fig. \ref{fig:procedure_seg} shows the procedure of the obstacle-aware road map segmentation method and the main intermediate results of an example. The main improvement lies in the obstacle entities. 
Geographic obstacle data usually stores in vector form in spatial databases (e.g., OpenStreetMap) and we convert the vector-based obstacle data into binary images. Each pixel represents a grid cell (`1' denotes the obstacle segments, and `0' stands for the background). Then, we fuse the obstacle raster with thinned road raster (after dilation and thinning), so that the obstacles entities will also be the boundaries for segmentation. The generated regions are called \textbf{atomic spatiotemporal elements} $\mathcal{P}^{\prime}$. Besides, we should use fine-grained level road data since we expect fine-grained spatial operation and the atomic spatial elements need to be as small as possible.

\subsubsection{Spatiotemporal Data Filtering}
The obstacle-aware road map segmentation method outputs $\mathcal{P}^{\prime}$, containing $N^{\prime}$ atomic spatial elements. Due to the heterogeneous spatial data density, many spatial atomic elements hardly contain spatiotemporal service data and are unnecessary to store. Filtering them helps reduce the scale of subsequent clustering problems and obtain better spatial granularity for supporting spatiotemporal services. We impose restrictions on the atomic elements based on the service data, filter out atomic elements whose data amount (e.g., daily average demand) is smaller than $\alpha$, and then get $N$ main atomic spatial elements.

\begin{figure*}[htbp]
  \centering
  \begin{minipage}[c]{0.63\linewidth}
  \includegraphics[width=1\linewidth]{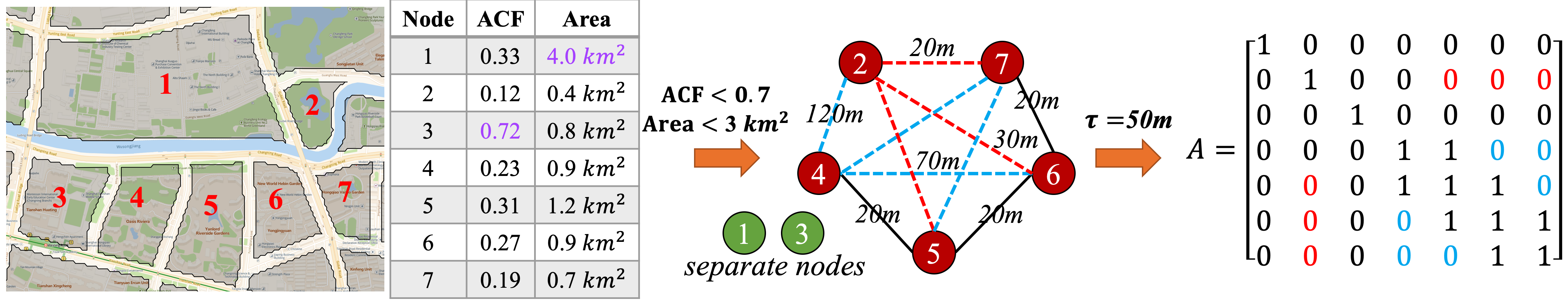}
  \caption{An illustrative example of the graph generation process. The nodes represent 7 atomic spatial elements and the adjacent matrix is calculated based on the geographic distance and the obstacle entities. The red and blue dotted lines denote that they can not be merged into the same cluster due to the river and far distance respectively.}
  \label{fig: graph_generation}
  \end{minipage}
  \hspace{2mm}
  \begin{minipage}[c]{0.35\linewidth}
  \includegraphics[width=1\linewidth]{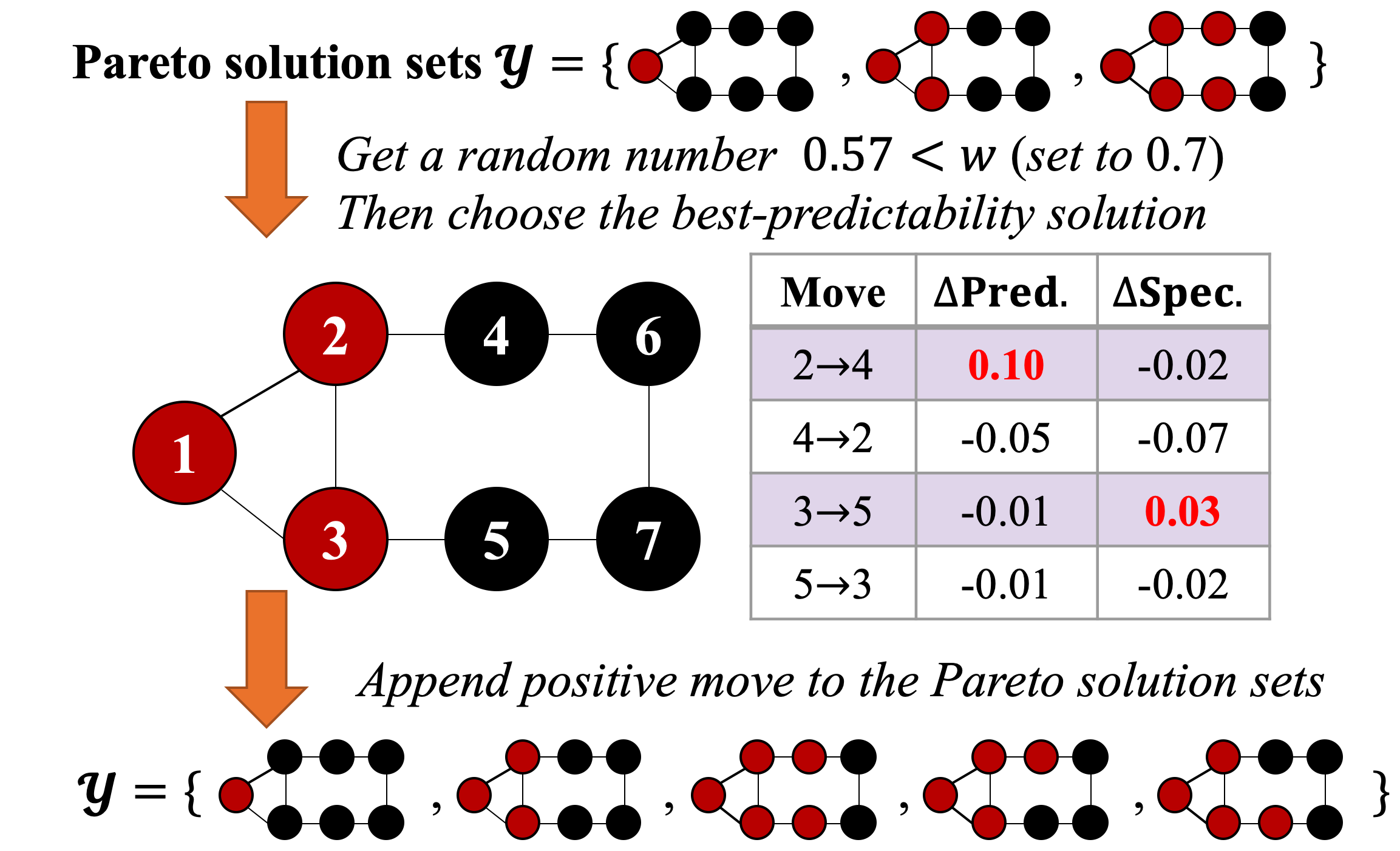}
  \vspace{-2em}
  \caption{An example of the predictability-specificity co-optimization process during an iteration. A$\rightarrow$B: append B to the cluster of A.}
  \label{fig: co_optimization}
  \end{minipage}
  \vspace{-1.3em}
\end{figure*}

\subsection{Atomic Spatial Element Clustering}
\subsubsection{Graph Generation} \label{adjacent_matrix}
To solve the clustering problem for region generation (Sec.~\ref{sub:problem_clustering}), we transform the problem into graph format so that the connected constraint (i.e. Eq. \ref{connected_constraints}) is easily satisfied by adopting many well-studied techniques (e.g., connected graph partition). Every atomic spatial element can be regarded as a node in the graph and their edges denote their `aggregatable' property. That is, the edges between two nodes represent that the predefined constraints (e.g., maximum area and geographic adjacent constraints) are still satisfied after merging these two connected nodes. In this section, we elaborate on building the edge sets of atomic spatial elements based on domain knowledge and geographic constraints.

\textbf{Domain Knowledge.} The atomic spatial elements may have the following properties, which indicate that the atomic elements cannot or unnecessarily aggregate with other atomic elements:

\begin{itemize}
    \item \textit{Oversize:} The road network may be scarce in some places (e.g., suburbs), and obstacle-aware road map segmentation may produce huge atomic elements whose area may exceed the maximum area specified by the service operator. At this time, aggregating the large atomic elements will generate an oversized region, which makes the operation difficult.
    \item \textit{Already Predictable:} In urban hot spots, such as commercial areas, small atomic elements may contain a lot of spatiotemporal data, and they may be predictable even if not aggregated with other atomic elements. In this situation, it is not necessary to aggregate, and small regions can retain fine-grained spatial granularity.
\end{itemize}

\label{geo_constraints}
\textbf{Geographic Constraints.} The `aggregatable' nodes must meet the following two geographic constraints (denote as $\mathcal{C}$), otherwise, the clustered regions are useless for spatiotemporal services:
\begin{itemize}
    \item \textit{Adjacent Constraint:} `Aggregable' atomic elements should be geographically adjacent. In practice, we usually set a small threshold $\tau$ (e.g., 50 m), and when the shortest distance between the two atomic elements is less than $\tau$, two atomic regions are defined as geographically adjacent.
    \item \textit{Obstacle Constraint:} Regions with good spatial semantics do not cross obstacle entities (e.g., rivers). To ensure that the aggregated regions still keep good spatial semantics, if there is an obstacle between the two atomic elements, there will be no edge between these two atomic spatial elements.
\end{itemize}

Fig. \ref{fig: graph_generation} shows an example of the graph generation process of 7 atomic spatial elements. First, based on the ACF and area attributes, Node 1 and Node 3 will be separate nodes due to oversize area and high predictability (i.e., ACF). We then calculate the geographic distance for the remaining 5 nodes (Node 2, 4, 5, 6, 7). Among these 5 nodes, there is a river between Node 2 and the remaining nodes, so there are no edges between them (marked with red dotted lines). Next, the distances between Node 2 and Node 4, Node 4 and Node 6, etc., exceed the given threshold ($\tau= 50m$) and therefore do not have edges between them (marked with blue dotted lines). Node 4 and Node 5, Node 5 and Node 6, etc., are closely adjacent and do not cross the river, and thus, can be aggregated (marked with black lines). With the above process, the graph generation module eventually produces a graph $G$ with atomic elements as nodes ($V$) and their adjacency matrix $A$ (equivalent to the edge set $E$ of $G$).

\subsubsection{Cluster Scale Estimation}\label{cluster_scale_estimation}

One challenge in solving the clustering problem is ensuring that the solutions meet the maximum area constraints (Ineq. \ref{area_constraints}). The maximum area constraint depends on the number of clusters ($M$) required to operate services in a city of a given size. For instance, if a $100\ km^2$ city has a maximum area of $5\ km^2$ for each region, 20 regions may be necessary; however, if this constraint changes to $10\ km^2$, only 10 regions are needed. To estimate the clustering scale and satisfy these constraints, we gradually increase $M$ and check whether they are met. We define an ideal minimum cluster number as $M^{*}$ which satisfies $\lceil \frac{\sum_{i=1}^Nts_i}{L} \rceil < M^{*} < N$. Here, $M^{*}$ represents the minimum cluster scale below which feasible aggregation results cannot be obtained.

To test whether maximum area constraints are satisfied at different trial cluster scales, we require a \textit{fast solver} that can obtain feasible aggregated solutions quickly. Existing balanced graph partition methods \cite{karypis1998fast,andreev2004balanced,sanders2013think} have been shown to be efficient enough for our purposes (a sparse graph with less than 10k nodes). We use \textit{Metis} software package as our \textit{fast solver} and assign node weights based on spatiotemporal data amounts while minimizing edge-cut tolerating up to 5\% unbalance. This method is called \textit{D-Balance}.

\subsubsection{Predictability-Specificity Co-optimization} \label{sec:co_solver}
The above \textit{fast-solver} can generate feasible but insufficiently superior solutions, and the specificity is not taken as the optimization objective. Hence, we design a heuristic predictability-specificity co-optimization algorithm that can iteratively refine multiple objectives, based on the famous node-swapping local search approaches (i.e., Fiduccia-Mattheyses \cite{FM_1980}). We summarize the proposed algorithm in Algorithm \ref{alg:local_search} (Appendix \ref{app:algorithm}).
Our algorithm first initializes Pareto solution sets $\mathcal{Y}$ with simple strategies (e.g., random growth or greedy). There are three initialization methods (details in Appendix \ref{sec:initial_methods}):
\begin{itemize}[leftmargin=0.24cm]
    \item \textit{D-Balance} balances the data across cluster (details in Sec.~\ref{cluster_scale_estimation}).
    \item \textit{Greedy} \cite{karypis1998fast} grows clusters by choosing the maximum gain.
    \item \textit{Fluid} \cite{pares2017fluid} aggregates nodes by random propagation.
\end{itemize}
Then, at each iteration, it first selects a candidate solution from $\mathcal{Y}$ and iteratively improves it by moving boundary nodes that connect two clusters to obtain the positive gain (obtaining better predictability or specificity) while not violating the geographic constraint $\mathcal{C}$. 

The way of selecting the candidate solutions from Pareto solution sets $\mathcal{Y}$ will determine the final solution quality. We hold the Pareto optimal solutions rather than weighting them into a single objective (e.g., linear scalarization \cite{miettinen2012nonlinear}). At the beginning of each iteration, we sample a random number $\in [0,1]$ from uniform distribution and compare it with a predefined parameter $w \in [0,1]$. If $w$ is bigger, we choose the best-predictability solution (probability $w$) from $\mathcal{Y}$. Otherwise, we select the best-specificity solution (probability 1-$w$). That is, $w$ represents the preference (probability) of optimizing the predictability objective. As the example shown in Fig.~\ref{fig: co_optimization}, we first choose the candidate solution with probability $w=0.7$. Then the selected best-predictability solution will serve as the starting solution for the refinement. We record all positive gain movements and append them into $\mathcal{Y}$ for the next iteration. Note that, in one iteration, even if we choose the best-predictability solution for refinement, we still record the better-specificity solutions during refinement. The algorithm stops when no more positive gain can obtain or achieve the max iterations. 

\textbf{Time Complexity}. The refinement process improves solutions by swapping the nodes having edges, and we will try every pair of nodes in the worst case. Since we limit the max iteration number to $n$, the time complexity of the co-optimization algorithm is $O(n\vert E\vert)$, where $E$ is the edge set of $G$.

\section{Implementation and Deployment}
\label{sub:deployment}
\begin{figure}[t]
    \centering
    \includegraphics[width=1\linewidth]{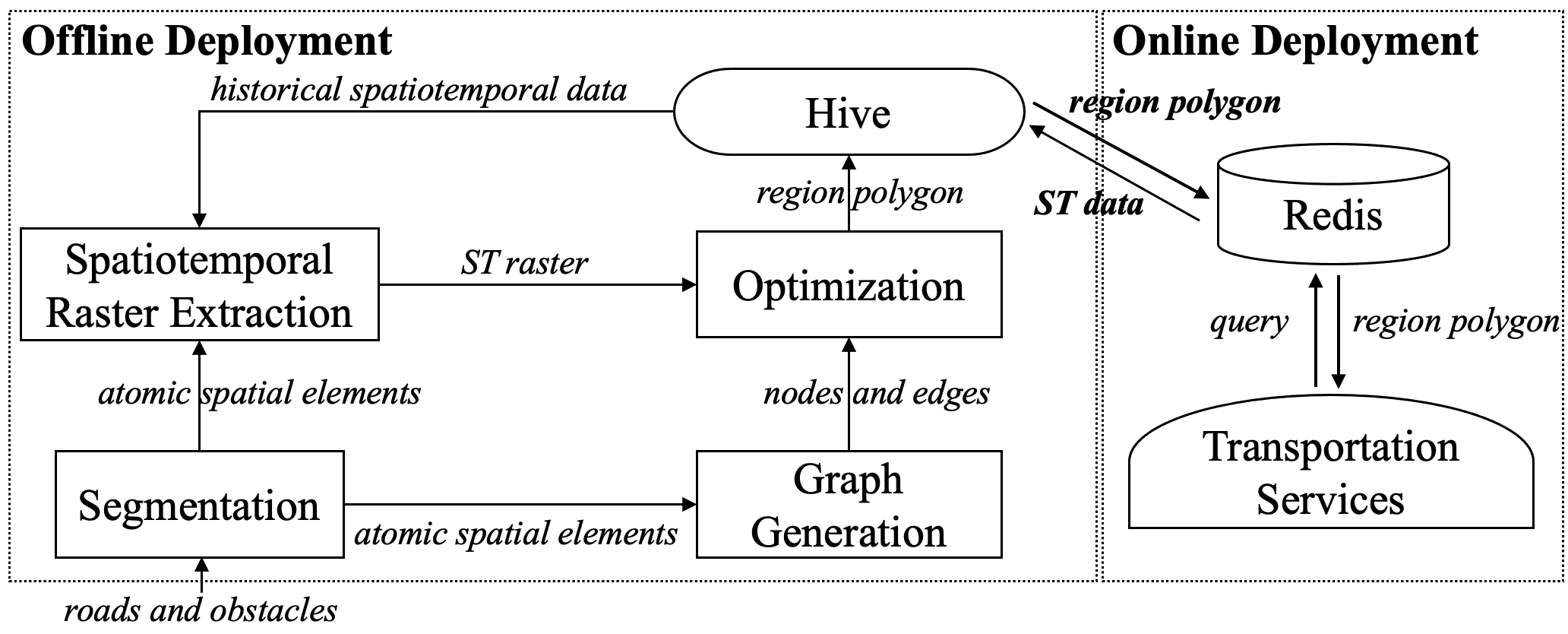}
    \vspace{-1em}
    \caption{The architecture of \textit{RegionGen} system.}
    \label{fig: deploy}
    \vspace{-1.em}
\end{figure}

As illustrated in Fig.~\ref{fig: deploy}, our system mainly has two parts, i.e., offline periodical region optimizing and online region query for the transportation services. In the offline optimizing part, we adopt the `T+1' mode to update the generated regions, which means the regions will be optimized and uploaded to Hive~\footnote{https://hive.apache.org/} in an offline manner on a daily basis and will be used for downstream spatiotemporal transportation services for the next day. In the online part, Redis\footnote{https://redis.io/}, an online real-time database, daily updates the region polygons from Hive and respond to online queries from online transportation services. For example, when providing demand heatmap services, region polygons are queried for visualization.

The offline part of \textit{RegionGen} is capable of optimizing fine-grained spatial atomic elements on a daily basis while the online transportation service only needs to look up the region polygons from Redis and the response time is around 100ms. The current run time of \textit{RegionGen} is around 2 hours (Sec.~\ref{sec:results}) with serial processing, which is already enough for `T+1' daily updates. Meanwhile, it can be accelerated by parallel processing. In each iteration, the co-optimizing algorithm in Sec. \ref{sec:co_solver} refines solutions by moving all boundary nodes, which can be parallelized (i.e., each processor deals with a part of nodes). In practice, our \textit{RegionGen} system has deployed to support online real-world services, such as demand heatmap visualization.

\section{Evaluation}
\label{sec:experiment}

\begin{table*}[t]
\small
\caption{Results on designated driver and freight transport dataset. M is the number of regions. The best two results are highlighted (\textbf{\textit{best}} is in bold and \textit{italic}, \textbf{second-best} is in bold). The MAPE@Recall results are given by \textit{STMGCN} \cite{geng2019spatiotemporal}. The specificity of \textit{DBSCAN} is not applicable (N/A) since it is a point clustering method.}
\vspace{-1.3em}
\setlength{\tabcolsep}{1.4mm}{
\begin{tabular}{lcccccccccccccc}
\toprule
& \multicolumn{4}{c}{\textbf{Designated Driver (M=913)}} & \multicolumn{3}{c}{\textbf{Freight Transport (M=724)}} \\ 
 \cmidrule(lr){2-5} \cmidrule(lr){6-9}
& ACF\_daily & Specificity & MAPE@97\% & Run Time & ACF\_daily & Specificity & MAPE@98\% 
& Run Time \\
\midrule
\multicolumn{3}{l}{\textbf{Baselines}} \\
\textit{Grid}  & 0.4186  & 0.3607 & 0.3647 & <1s & \textbf{0.3596} & 0.1701 & \textbf{0.3634} & <1s \\
\textit{Hexagon} & 0.4307  & 0.3781 & 0.3615 & <1s & 0.3579 & 0.1853 & 0.3657 & <1s \\
\textit{DBSCAN} & 0.3998  & N/A & 0.3711  & $\sim$7mins & 0.3189  & N/A & 0.4092 & $\sim$4mins \\
\textit{BSC} & 0.4421  & 0.3791 & 0.3580 & $\sim$4mins & 0.3378 & 0.1879 & 0.3852 & $\sim$5mins \\
\textit{GCSC} & 0.4240  & 0.3615 & 0.3639 & $\sim$5mins & 0.3295 & 0.1812 & 0.3966 & $\sim$3mins \\
\midrule
\textit{RegionGen (best specificity)} & \textbf{0.4577} & \textbf{\textit{0.3902}} & \textbf{0.3501} & $\sim$100mins & 0.3497  & \textbf{\textit{0.2535}} & 0.3698 & $\sim$50mins\\
\textit{RegionGen (best ACF)}  &  \textbf{\textit{0.4740}}  & \textbf{0.3851} & \textbf{\textit{0.3321}} & $\sim$100mins &  \textbf{\textit{0.3719}}  & \textbf{0.2502} & \textbf{\textit{0.3305}} & $\sim$50mins \\
\bottomrule
\end{tabular}}
\label{table: DD_FT_results}
\vspace{-1.2em}
\end{table*}

\subsection{Datasets, Baselines, and Experiment Settings}
We collect three transportation service datasets (designated driver, freight transport, and taxi demand) as well as the geographic entities (roads and obstacles). Dataset details are in Appendix~\ref{data_description}.
We implement widely-adopted manually-specified region generation methods (\textit{Grid}, \textit{Hexagon}) and data-driven baselines (\textit{DBSCAN} \cite{dbscan_1996}, \textit{GSC} \cite{li_traffic_2015}, \textit{GCSC} \cite{chen_dynamic_2016}). Our experiment platform is a server with 10 CPU cores (Intel Xeon CPU E5-2630), and 45 GB RAM. 
More baseline and experimental setting details are in Appendix~\ref{setting_and_baselines}.

\subsection{Evaluating with Spatiotemporal Prediction}
To evaluate the effectiveness of \textit{RegionGen} for accurately operating spatiotemporal services, we conduct demand predictions on three spatiotemporal service datasets, which are the basic capabilities of various downstream tasks, such as scheduling idle drivers. Specifically, we predict the demand in the next hour for all datasets.

\subsection{Evaluation Metric}
We evaluate the quality of the generated region by two operational characteristics, namely \textbf{ACF\_daily} and service \textbf{specificity}. 
Besides, popular metrics (including RMSE and MAE) are not feasible to directly evaluate the prediction performance, since the prediction ground truths of various region generation methods are different. The Mean Absolute Percentage Error (\textbf{MAPE}) measures the relative errors and thus different predictive objects are comparable. However, different regions contain varying amounts of service data, to make a fair comparison, we ensure the amount of service data of different regions is approximately equal (i.e., demand recall). Specifically, we filter out the regions whose average daily demand is less than 1, and get recalls of 97\%, 98\%, and 99.8\%, respectively, in the designated driver, freight transport, and taxi demand datasets.

\subsection{Results and Analysis} \label{sec:results}
\subsubsection{Main results}
In Table~\ref{table: DD_FT_results}, we report the ACF\_daily, Specificity, and MAPE@Recall on the designated driver and freight transport dataset under the same clustering scale. \textit{RegionGen} gets Pareto optimal solutions ($w$ is 0.7) and we report two solutions best in the ACF\_daily and specificity metrics respectively. 
In Table \ref{table:results_taxi}, we report the results of ACF\_daily and MAPE@Recall on the taxi dataset, since the latitude and longitude of the original data are aggregated into the census tracts, making the `specificity' metric inapplicable. Table~\ref{table: DD_FT_results} shows that \textit{RegionGen} achieves better ACF\_daily than the baselines and gets corresponding lower MAPE@Recall. Especially, \textit{RegionGen (Best ACF)} consistently outperforms baselines in terms of MAPE@Recall, decreasing 3.2\% and 3.3\% than \textit{Grid} in the designated driver and freight transport dataset. The above observations demonstrate the generated regions with better predictability support operating more accurate services. They also provide us with new insight that, besides predictive models, the prediction performance can be significantly improved by generating regions with better predictability. The results on the taxi dataset in Table~\ref{table:results_taxi} are similar. \textit{RegionGen} consistently gets the best ACF\_daily and prediction accuracy. In addition, we record the run time of \textit{RegionGen}. For one city, \textit{RegionGen} can finish region generation in two hours, which is enough for real-life deployment and usage (detailed discussion in Sec.~\ref{sub:deployment}).

\subsubsection{Robustness analysis on open datasets}
The main results are conducted on two private spatiotemporal service datasets.
To help reproduce our results, we also experiment on an open dataset, Chicago taxi demand.
Moreover, we conduct spatiotemporal prediction based on various state-of-the-art models, including \textit{STMGCN} \cite{geng2019spatiotemporal}, \textit{GraphWaveNet} \cite{graphwavenet_2019}, \textit{GMAN} \cite{GMAN_2020} and \textit{STMeta} \cite{STMeta}, as these four models have performed competitively in a recent benchmark study \cite{STMeta}.
Results in Table \ref{table:results_taxi} show that \textit{RegionGen} outperforms baselines consistently.
Despite small differences in the prediction accuracy of the four models, MAPE is highly correlated to ACF\_daily. For instance, \textit{RegionGen} obtains the highest ACF\_daily, and achieves the lowest MAPE consistently under four models. 
This confirms that our choice of ACF\_daily as a measurement for the predictability of regions is acceptable and effective. 

\begin{table}[h]
\small
\caption{Results on taxi demand dataset. M is the number of regions. The best two results are highlighted (\textbf{\textit{best}} is in bold and \textit{italic}, \textbf{second-best} is in bold). The original dataset stores the location at the census-tract level, so "Specificity" cannot be computed; \textit{RegionGen} is set to optimize ACF only ($w$=1).}
\setlength{\tabcolsep}{1.3mm}{
\begin{tabular}{lccccc}
\toprule
& \multicolumn{5}{c}{\textbf{Taxi Demand (M=95)}} \\ 
\cmidrule(lr){2-6} 
& \multirow{2}{*}{ACF\_daily}  & \multicolumn{4}{c}{MAPE@99.8\%}\\
\cmidrule(lr){3-6}
& & \textit{STMGCN} & \textit{STMeta} & \textit{GraphWaveNet}& \textit{GMAN}\\
\midrule
\multicolumn{2}{l}{\textbf{Baselines}}  \\
\textit{Grid}  & 0.3018  & 0.4152 & 0.4050 & 0.3904&0.3730\\
\textit{Hexagon} & 0.3076 & 0.4094 & 0.4016 & 0.3855&0.3700  \\
\textit{DBSCAN}& 0.3323   & 0.4011 & 0.3921 & 0.3776&0.3668 \\
\textit{BSC} & \textbf{0.3691} & \textbf{0.3526} & \textbf{0.3456} & \textbf{0.3308}&\textbf{0.3434} \\
\textit{GCSC} & 0.3402 & 0.3769 & 0.3674 & 0.3494&0.3561 \\
\midrule
\textit{RegionGen}  & \textbf{\textit{0.3841}}  & \textbf{\textit{0.3409}} & \textbf{\textit{0.3295}} & \textbf{\textit{0.3123}}&\textbf{\textit{0.3272}} \\
\bottomrule
\end{tabular}}
\label{table:results_taxi}
\vspace{-1.4em}
\end{table}

\newcommand{\fourfigcol}{-2mm}
\begin{figure*}[htbp]
\centering
\subfigure[]{
\includegraphics[width=0.24\linewidth]{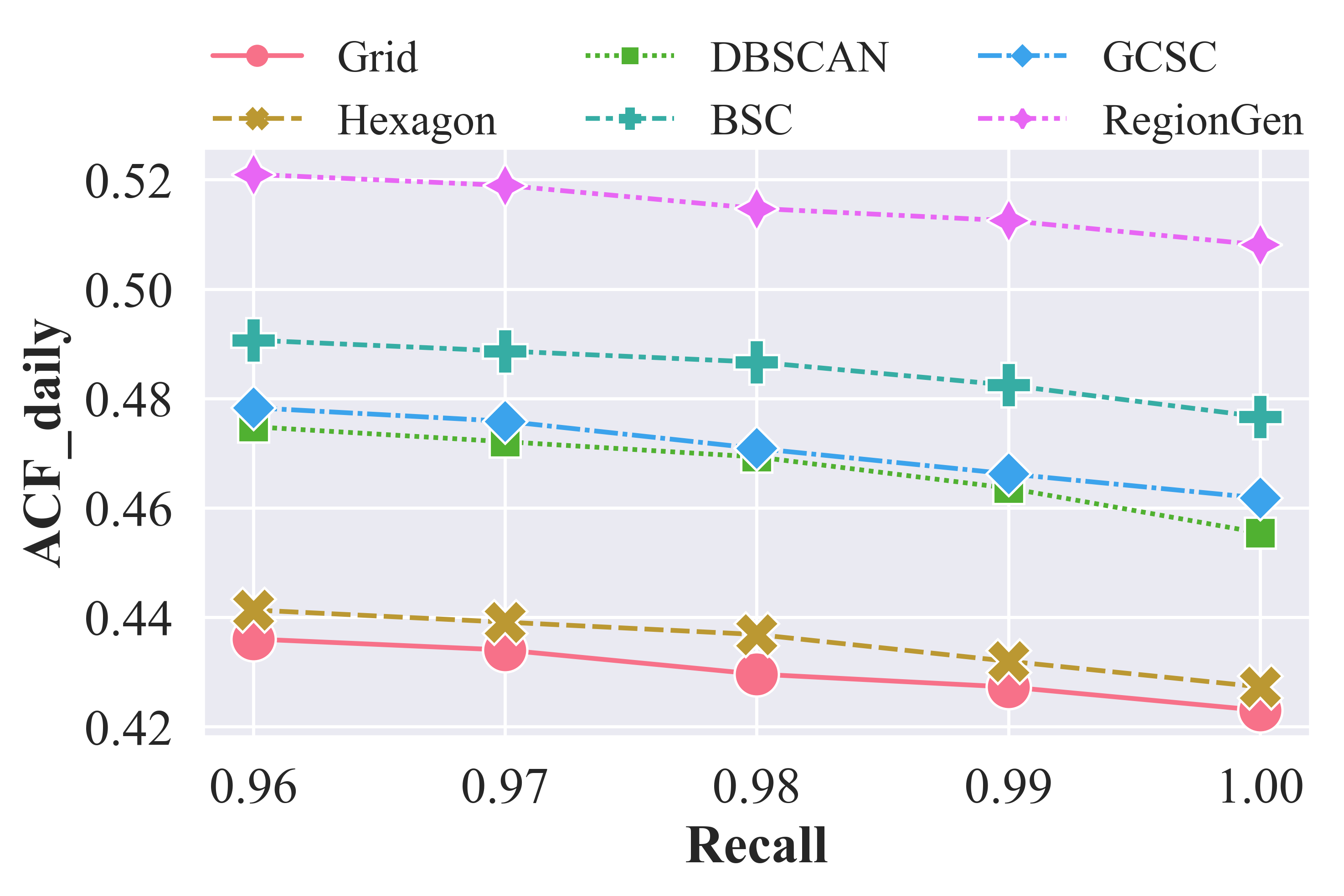}
\label{fig:acf_recall}
}\hspace{\fourfigcol}
\subfigure[]{
\includegraphics[width=0.24\linewidth]{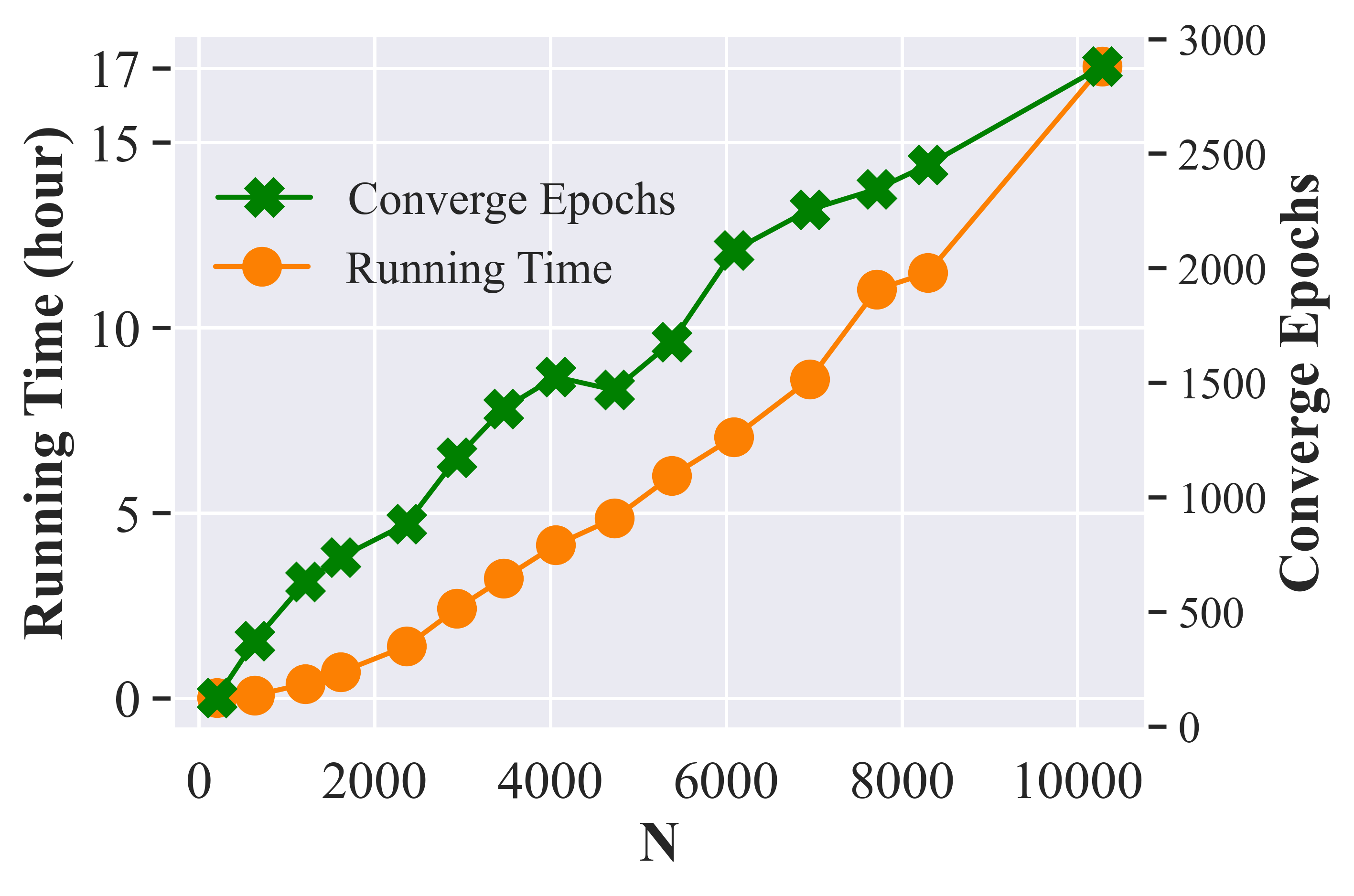}
\label{fig:scalability}
}\hspace{\fourfigcol} 
\subfigure[]{
\includegraphics[width=0.24\linewidth]{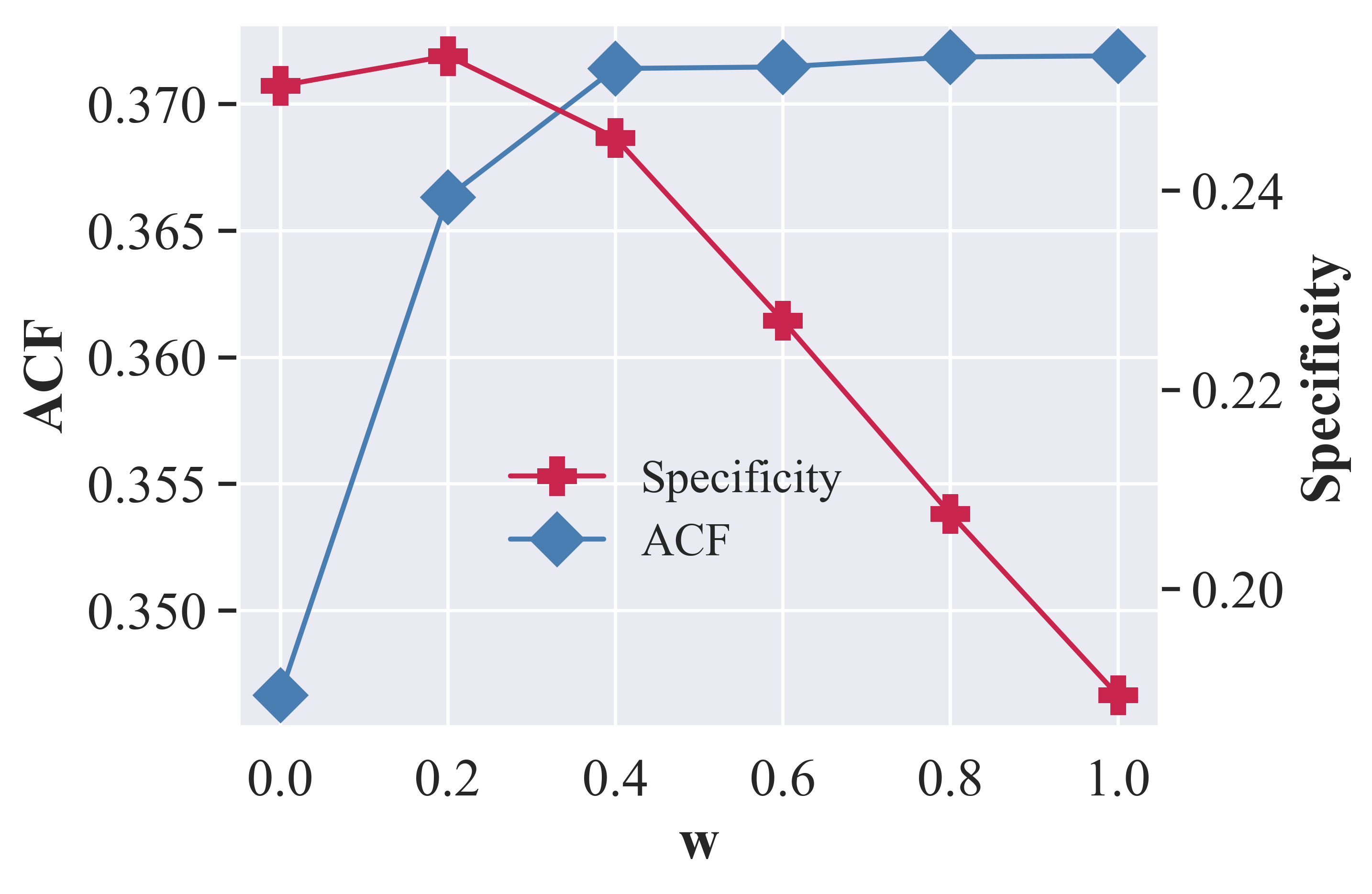}
\label{fig:weight_acf}
}\hspace{\fourfigcol} 
\subfigure[]{
\includegraphics[width=0.24\linewidth]{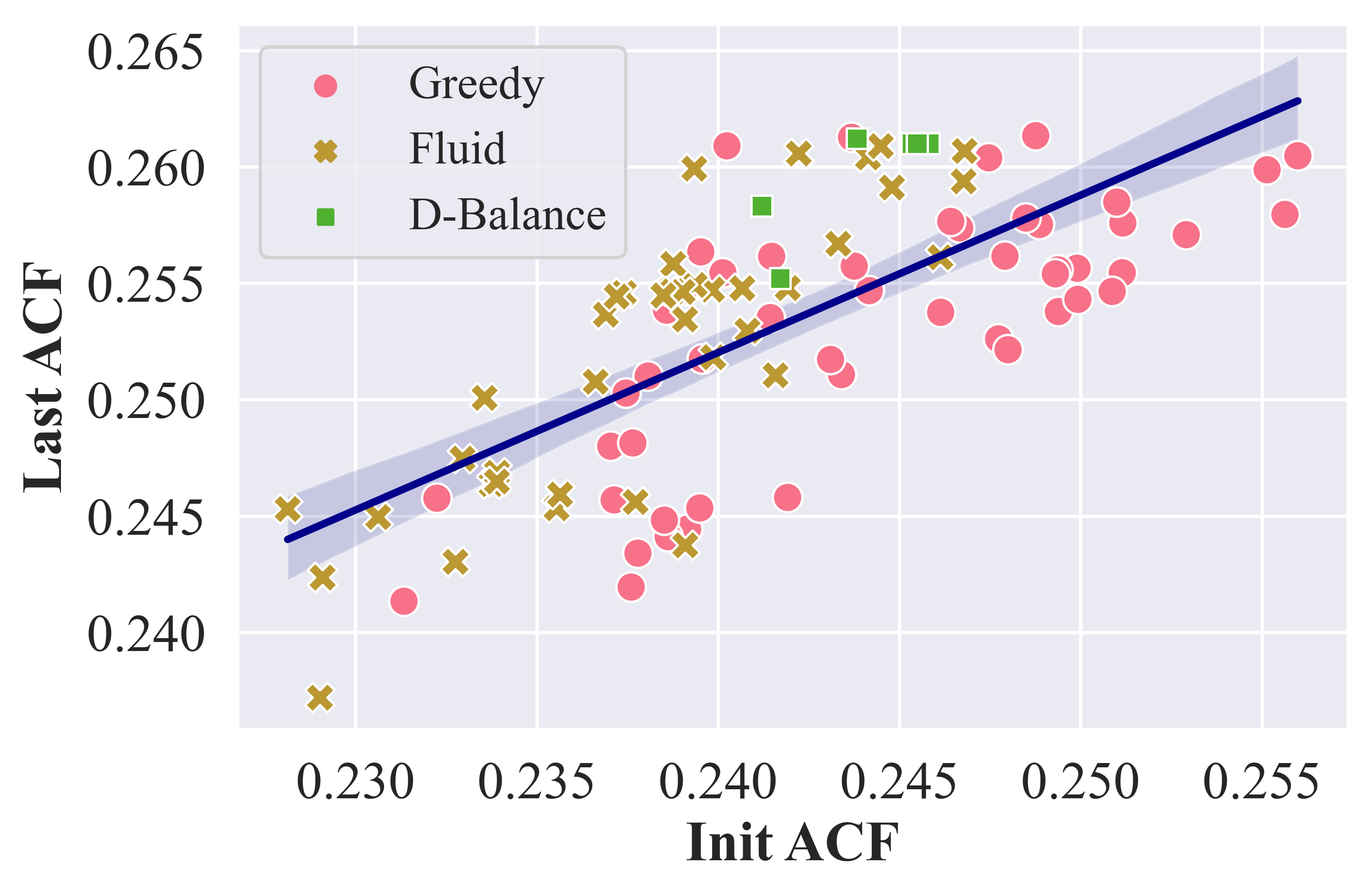}
\label{fig:init_comparison}
}
\vspace{-1.4em}
\caption{(a) ACF\_daily vs. recall; (b) Speed vs. the Scalability; (c) Performace vs. Parameter $w$; (d) Effect of initialization methods}
\vspace{-1.4em}
\end{figure*}

\subsubsection{Robustness analysis under different recall settings}
To see whether \textit{RegionGen} is robust under different recall settings, we unceasingly remove the tailed atomic spatial elements that are with the fewest data samples. Fig.~\ref{fig:acf_recall} show how ACF\_daily changes by varying the recall. We observe that \textit{RegionGen} (marked with red lines) consistently achieves the best ACF\_daily, demonstrating its robustness. 
Among baselines, with the recall decreasing, ACF\_daily of \textit{DBSCAN} increases the fastest (green lines). It shows that \textit{DBSCAN} has a relatively obvious long-tail phenomenon; that is, the ACF\_daily of the tailed elements is much lower than the other elements, which may incur inconvenience for operating services upon these tailed regions. 
Besides, fixed-shape methods, \textit{Grid} and \textit{Hexagon} (blue lines) perform consistently the worst.
Note that fixed-shape regions are still popular for spatiotemporal service management in practice, but these results again point out their weakness. 
Hence, it would be expected that new region generation technologies, such as \textit{RegionGen}, will significantly advance the field.

\subsubsection{Analysis of scalability} \label{scalability_analysis}
To analyze the scalability of \textit{RegionGen}, we conduct experiments on different scales (i.e., the number of atomic elements $N$). 
Fig. \ref{fig:scalability} shows our results. We explored the change of \textit{RegionGen} with the scale by dividing the target area into different numbers of atomic elements (from 100 to 10,000).\footnote{In this experiment, we choose grids as the atomic spatial elements since they easily adapt to different granularity.} We observe that as the scale increases, the running time and the converge epochs of the algorithm also increase synchronously. It is worth noting that 10,000 atomic elements can already support fine-grained spatiotemporal service in a metropolis like Shanghai (i.e.,  each atomic element covers around $0.01$ km$^2$); with 10,000 atomic elements, \textit{RegionGen} takes 17 hours, which still satisfies the need for the `T+1' update mode in our realistic deployment (Sec.~\ref{sub:deployment}). This demonstrates that \textit{RegionGen} is capable of optimizing very fine-grained spatial atomic elements.

\subsubsection{Effect of $w$}
In Fig. \ref{fig:weight_acf}, we display the ACF metric of the best-predictability solution and the Specificity metric of the best-specificity solutions from different Pareto solution sets obtained by changing $w$, the probability of optimizing the predictability objective. We observe that: (i) with the increase of $w$, the ACF metric gradually increases and the specificity metric decreases; (ii) when $w$ is set to 0.4, the $w$-hold mechanism prefers to optimize specificity, but still achieves solutions with reasonable predictability. Hence, in practice, we may obtain a solution with both good predictability and high specificity by setting an appropriate $w$.

\newcommand{\subfigcol}{-2mm}
\begin{figure*}[t]
\begin{minipage}[c]{0.75\linewidth}
\centering
\subfigure[Heatmap]{
\includegraphics[width=0.134\linewidth]{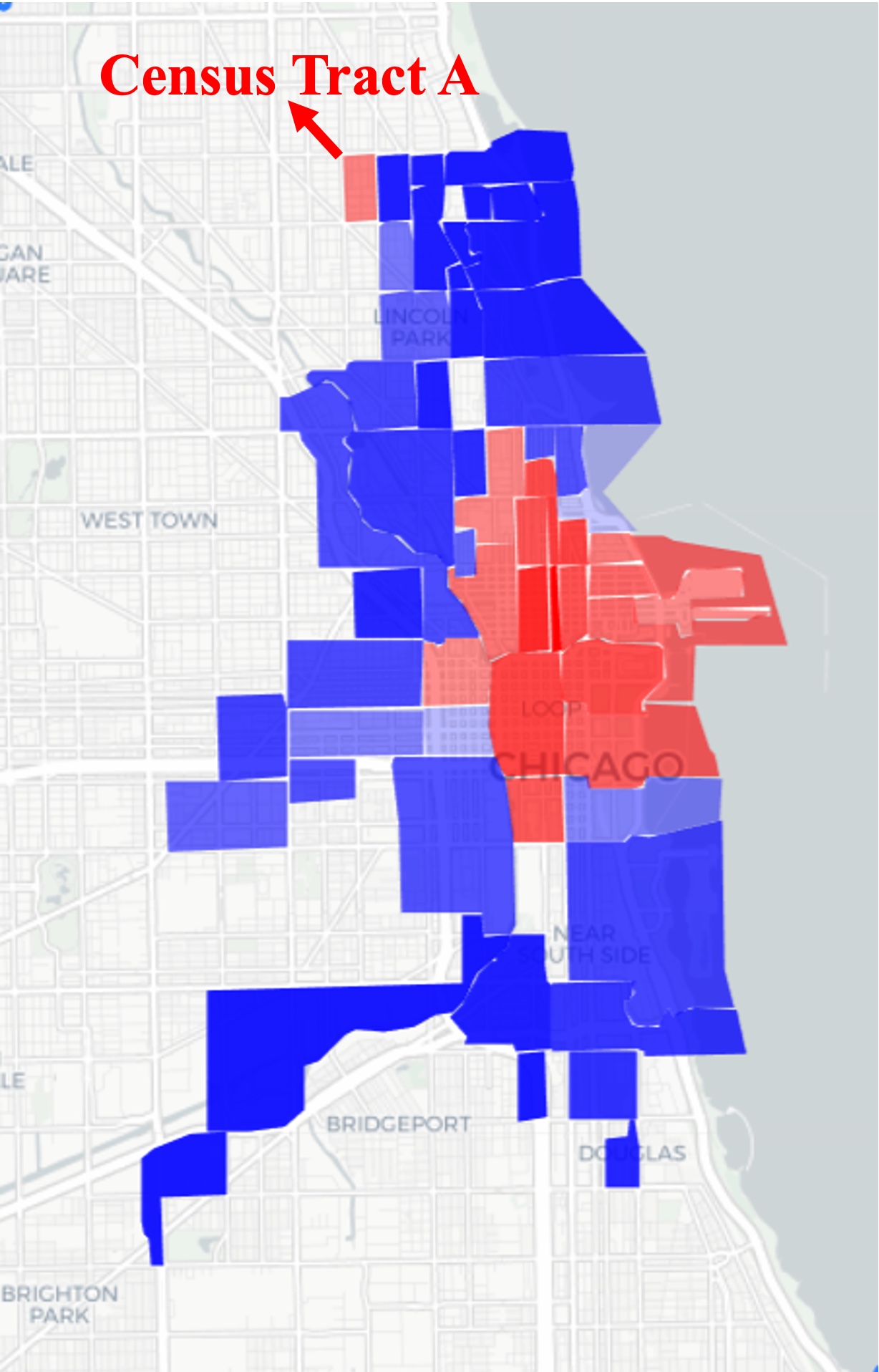}
    \label{vis_heatmap}
}\hspace{\subfigcol}
\subfigure[Grid]{
\includegraphics[width=0.134\linewidth]{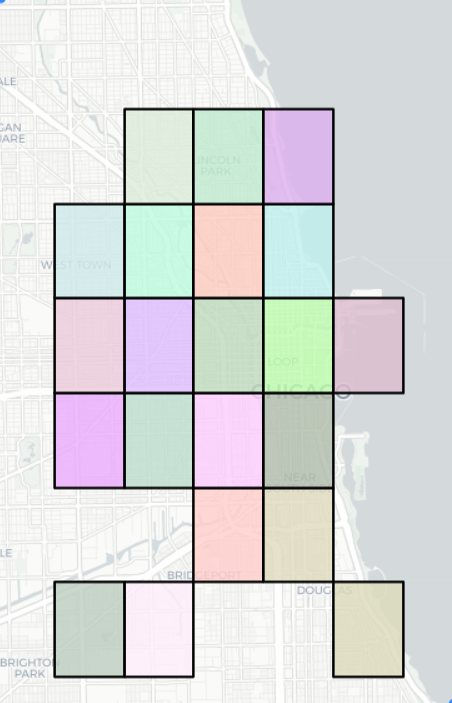}
    \label{vis_grid}
}\hspace{\subfigcol}
\subfigure[Hexagon]{
\includegraphics[width=0.134\linewidth]{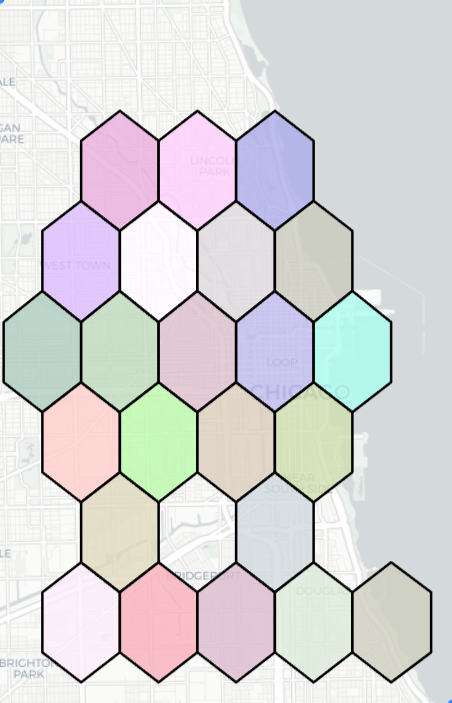}
    \label{vis_hexagon}
}\hspace{\subfigcol}
\subfigure[DBSCAN]{
\includegraphics[width=0.134\linewidth]{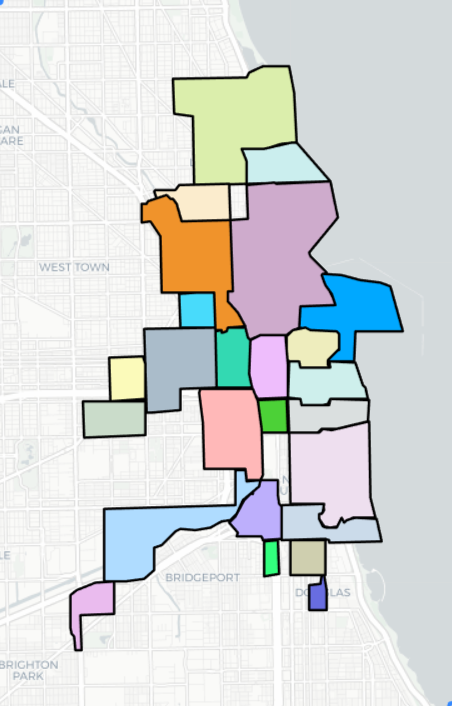}
    \label{vis_dbscan}
}\hspace{\subfigcol}
\subfigure[BSC]{
\includegraphics[width=0.134\linewidth]{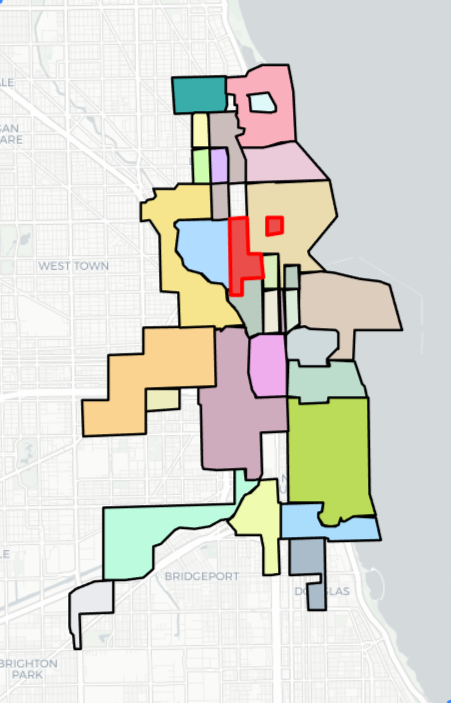}
    \label{vis_bsc}
}\hspace{\subfigcol}
\subfigure[GCSC]{
\includegraphics[width=0.134\linewidth]{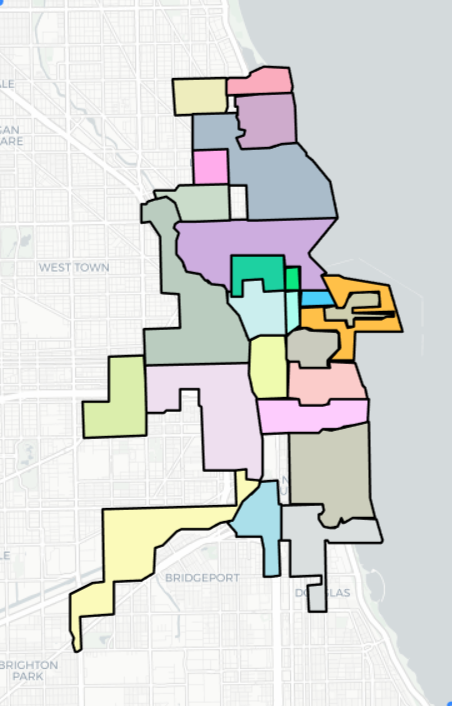}
    \label{vis_gcsc}
}\hspace{\subfigcol}
\subfigure[RegionGen]{
\includegraphics[width=0.134\linewidth]{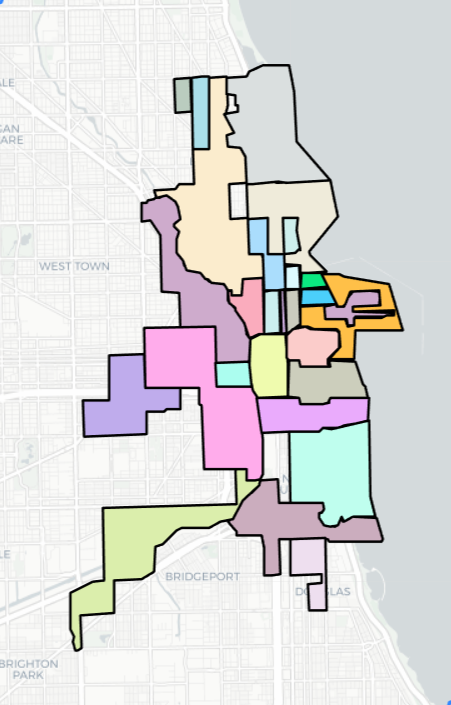}
    \label{vis_localsearch}
}\hspace{\subfigcol}
\vspace{-.8em}
\caption{Region clusters in Chicago generated by the baselines and RegionGen.}
\label{vis}
\end{minipage}
\begin{minipage}[c]{0.245\linewidth}
\includegraphics[width=1\linewidth]{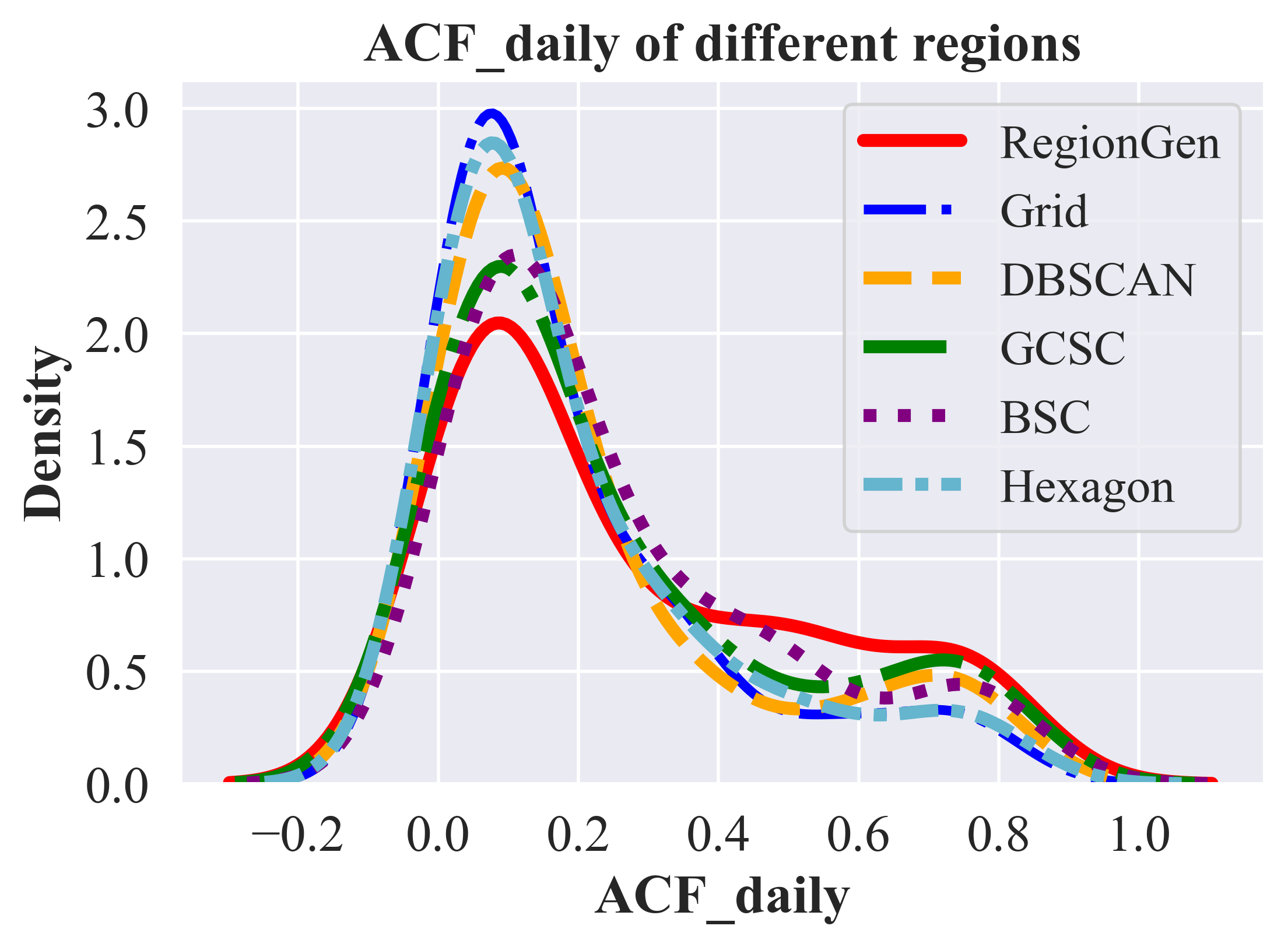}
\caption{ACF distribution.}
\label{fig: demand_dis}
\end{minipage}
\vspace{-.8em}
\end{figure*}

\subsubsection{Effect of different initialization methods}
In Fig.~\ref{fig:init_comparison}, we compare different initialization methods including \textit{D-Balance}, \textit{Greedy}, and \textit{Fluid} by generating initial solutions with 30 different random seeds for each method. We record the initial ACF (called Init ACF) and the ACF when the algorithm stops (called Last ACF). We observe that (i) the final solutions given by \textit{Greedy} and \textit{Fluid} are quite different with diverse random seeds (the difference in terms of ACF exceeds 0.02); (ii) the Init ACF is linearly related to the last ACF, which inspires us to choose an initial solution with better quality for further optimization; (iii) The Last ACF of \textit{D-Balance} solutions are better than \textit{Greedy} and \textit{Fluid} (closer to the upper area), demonstrating that it is more capable and more robust to be the initialization method for generating high-quality solutions.

\subsection{Case Study}
We visualize the generated regions of baselines and \textit{RegionGen} in the downtown area of Chicago on the taxi demand dataset. We filter out regions with few historical service data for all region generation methods and each color represents a region. Fig. \ref{vis_heatmap} displays the demand heatmap and red indicates high-density areas.
In Fig. \ref{vis_grid} and Fig. \ref{vis_hexagon}, \textit{Grid} and \textit{Hexagon} generate regions with poor spatial semantics, while hotspots and cold areas are of the same spatial granularity. 
Based on census tracts, in Fig. \ref{vis_dbscan}, regions clustered by \textit{DBSCAN} are with good spatial semantic meaning. \textit{DBSCAN} prefers to aggregate more census tracts in the high-density area (i.e., hotspots with many data points), which offers excellent predictability but low spatial granularity. At the same time, in cold areas with few data points, \textit{DBSCAN} preserves a single census tract (good spatial granularity) but with bad predictability. 
In Fig. \ref{vis_bsc}, the aggregate regions by \textit{BSC} will not be oversize like \textit{DBSCAN} since the first partition stage in \textit{BSC} makes all aggregated regions geographically close. However, the nearby census tracts may not be strictly adjacent, and nonadjacent census tracts with similar demand matrices will also be aggregated (marked with red boxes), resulting in inappropriate clustering results. 
In Fig. \ref{vis_gcsc}, despite generating spatial continuous regions and obtaining sufficient adaptive granularity (small regions in hotspots and big regions in cold areas), \textit{GCSC} may still fall short of successfully optimizing the predictability. For example, census tract A is predictable (with much data to support clear daily regularity), yet \textit{GCSC} continues to aggregate it with other census tracts.
In Fig. \ref{vis_localsearch}, \textit{RegionGen} gets reasonable adaptive granularity and good predictability (small regions in hotspots and big regions in cold areas).
Specifically, \textit{RegionGen} takes census tract A (already predictable) alone as a cluster, demonstrating that it optimizes the predictability well.

In Fig. \ref{fig: demand_dis}, we explore ACF\_daily distribution of different regions on baselines and \textit{RegionGen}. For \textit{RegionGen} (red line), there are fewer regions with low ACF\_daily than all baselines, while having more regions with larger ACF\_daily. Therefore, \textit{RegionGen} obtains better predictability by balancing the spatiotemporal data over different regions. That is, it prefers to cluster fewer atomic spatial elements in hotspots and more in cold areas.

\section{Related Work}
With the wide adoption of GPS-equipped devices and the great success in machine learning models, massive historical spatiotemporal data (e.g., GPS trajectory data \cite{VRE_2022_vldb}) has been widely used to support intelligent transportation services, including traffic prediction \cite{shao_traffic_2022,modeling_network_flow_2022,multi_step_relation_traffic_2022,pretrain_traffic_2022,dynamic_multi_traffic_2021,network_imputation_2021,ode_traffic_2021,quantify_uncertainty_2021,stnorm_2021,trajNet_2021,hybrid_navigation_2020}, travel time estimation \cite{ssml_route_time_2021,yuan_travel_time_2020,bustr_bus_time_2020,contextual_travel_time_2020,deeptrans_2020}, transportation route recommendation \cite{liu_route_recomm_2021,polestar_2020,hydra_route_2019}, trajectory similarity computing and outlier detection \cite{TrajGAT_2022,ST_traj_2022,liu_icde_2020,deepTEA_2022}, bus route planning \cite{bus_route_planning_2021,gan_route_planning_2021,fast_traj_clustering_2019}, outdoor advertising \cite{outdoor_advertising_2019}, and crowdsourcing \cite{experiment_crowdsourcing_2018,price_crowdsourcing_2018}. The transportation services may be operated upon specified areal units, e.g. by fixed-size grids \cite{selective_traffic_2022,yuan_demand_2021,fang_multi_source_2021,traj_local_differential_2021,graph_traj_similarity_2021,curbGan_2020,dynamic_attention_2020,co_prediction_2019,meta_learning_traffic_2019}. Existing transportation services may benefit from our region generation framework. For example, for spatial crowdsourcing pricing applications (e.g., food delivery services), previous research demonstrated that spatial crowdsourcing needs to price for multiple local markets based on the spatiotemporal distributions of tasks and workers than seek a unified optimal price for a single global market \cite{price_crowdsourcing_2018}. The regions created by our framework are more suitable for estimating the spatiotemporal distribution of works (e.g., making the prediction of the supply of workers more accurate) and thus the pricing strategies are easier to give than grids \cite{experiment_crowdsourcing_2018,
price_crowdsourcing_2018}. Moreover, with better spatial semantic meaning, our regions may support further analysis correlating with regions' functionality.

\section{Conclusion}
In this paper, we present a unified data-driven region generation framework, called \textit{RegionGen}, which can flexibly adapt to various operation requirements of spatiotemporal services while keeping spatial semantic meaning. 
To keep the good spatial semantic meaning, \textit{RegionGen} first segments the whole city into atomic spatial elements based on the fine-grained road networks and obstacle entities (e.g., rivers). 
Then, it aggregates the atomic spatial elements by maximizing key operating characteristics such as predictability and specificity. Extensive experiments have been conducted in three transportation datasets including two industrial datasets and an open dataset. The results demonstrate that \textit{RegionGen} can generate regions with better operating characteristics (including spatial semantic meaning, predictability, and specificity) compared to other region generation baselines under the same granularity. 
Moreover, we conduct demand prediction services upon the generated regions, and \textit{RegionGen} still achieves the best performance, verifying its effectiveness. 
As a pioneering study toward the adaptive region generation problem, we expect that our research can benefit online transportation platforms to provide intelligent and satisfactory transportation services.

\begin{acks}
This work was partly supported by National Science Foundation of China (NSFC) Grant No. 61972008 and CCF-DiDi GAIA Collaborative Research Funds for Young Scholars. Yongxin Tong's work was partially supported by National Science Foundation of China (NSFC) Grant No. U21A20516.
\end{acks}

\bibliographystyle{ACM-Reference-Format}
\balance
\bibliography{sample-base}

\clearpage
\appendix

\section{Illustrative Evidences}

\subsection{ACF\_daily is a proper proxy for measuring predictability}
\label{app:acf}

\begin{figure}[h]
  \centering
  \includegraphics[width=0.7\linewidth]{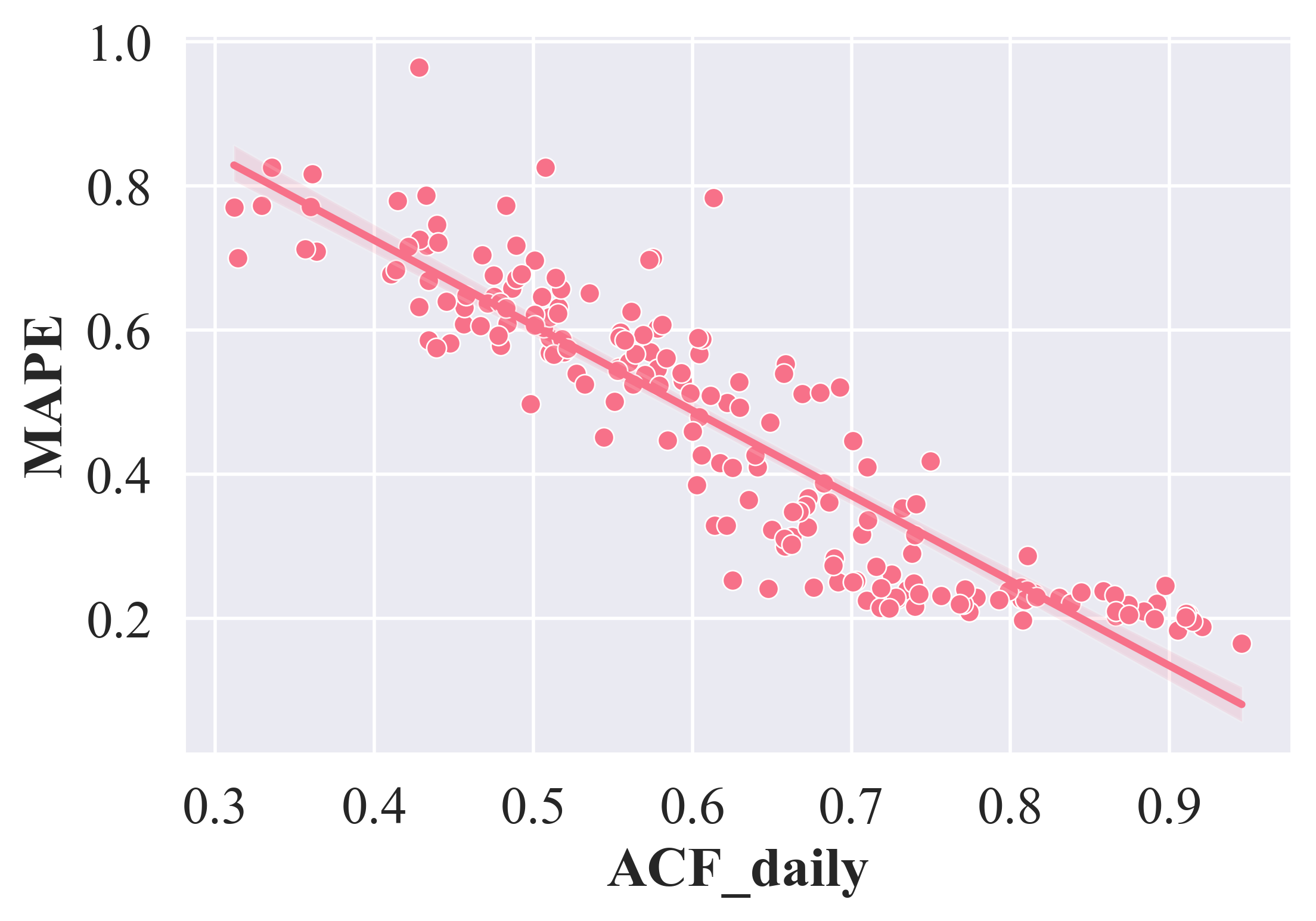}
  \caption{The ACF of different time series and the corresponding prediction error (Mean Absolute Percentage Error) given by STMGCN \cite{geng2019spatiotemporal}. The prediction is conducted on the freight transport dataset (Sec. \ref{sec:experiment}) for one-hour time slots. The ACF is computed at the daily scale, i.e., $k$ is set to 24 in Eq.~\eqref{eq:acf}.}
  \label{fig:acf_predictability}
  \vspace{-1em}
\end{figure}

\subsection{More data samples support more obvious daily regularity}
\label{app:more_data}

\begin{figure}[htbp]
\centering
\includegraphics[width=0.7\linewidth]{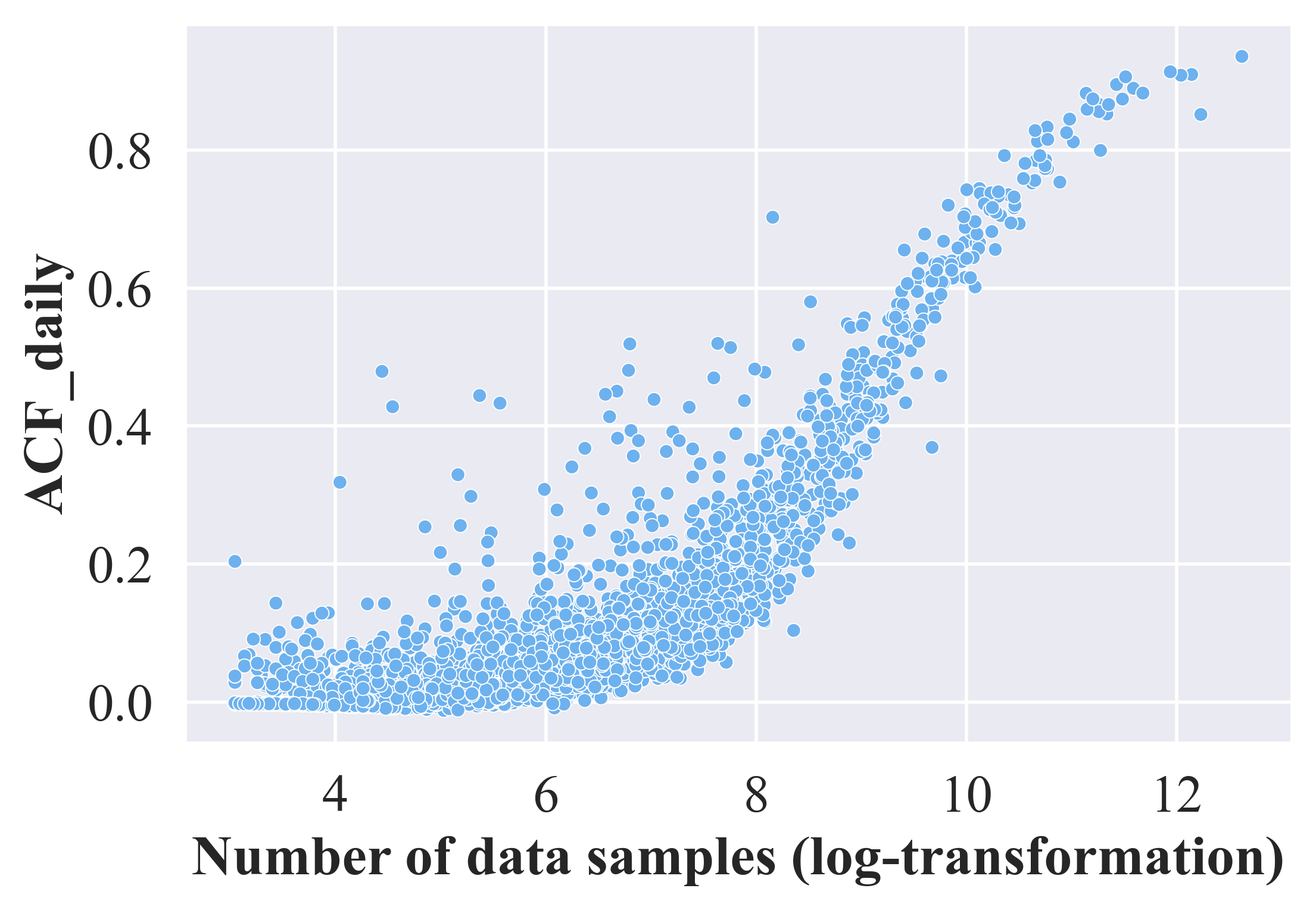}
\caption{More data samples support more obvious daily regularity, i.e., ACF\_daily. Every data point represents an atomic spatial element in the freight transport dataset.}
\label{fig: data_acf}
\vspace{-1.5em}
\end{figure}

\subsection{Example of calculating specificity}
\label{app:example_spec}

\begin{figure}[h]
    \centering
    \subfigure[The serviced area]{
    \includegraphics[width=0.47\linewidth]{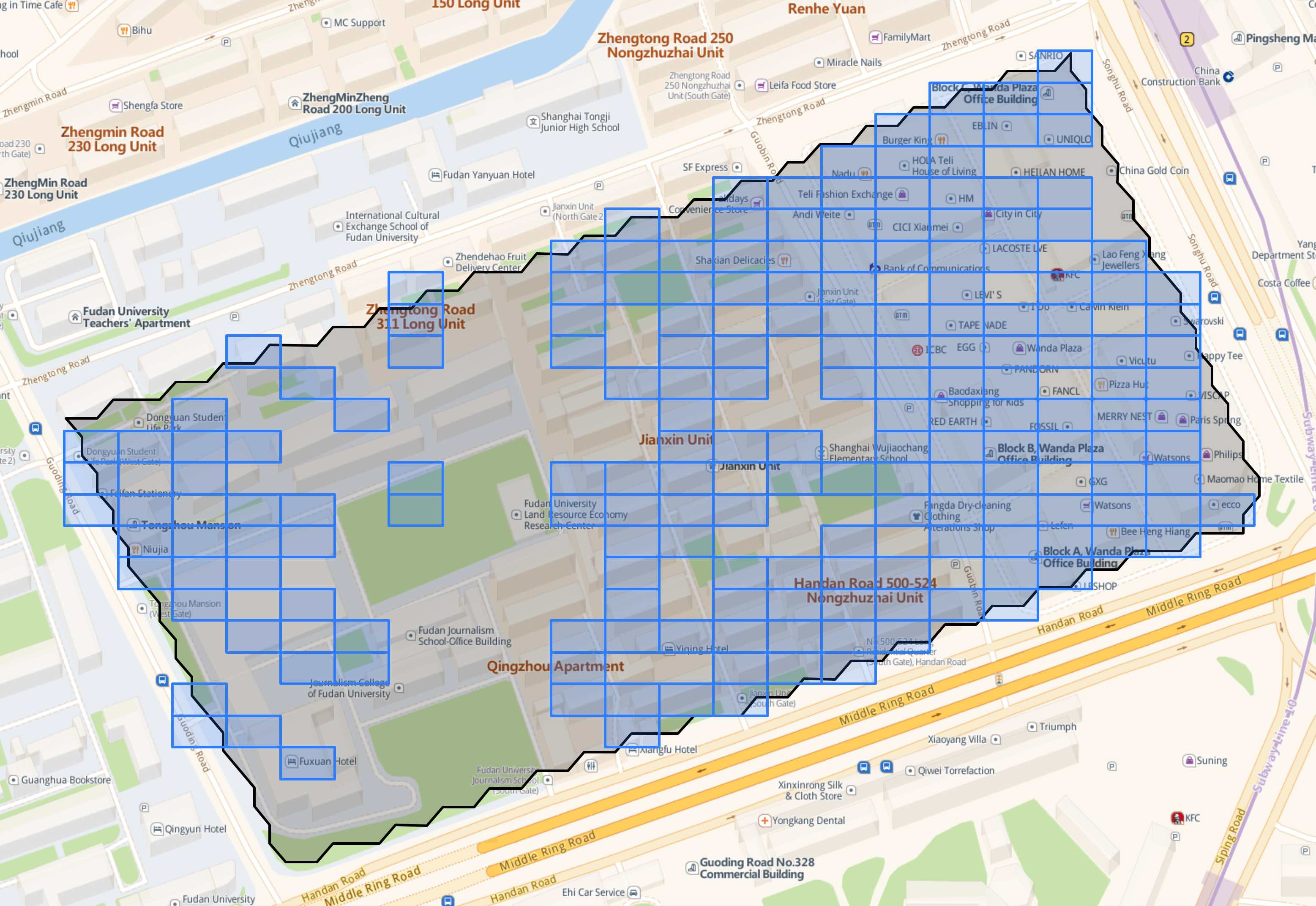}
    }
    \subfigure[The total area]{
    \includegraphics[width=0.47\linewidth]{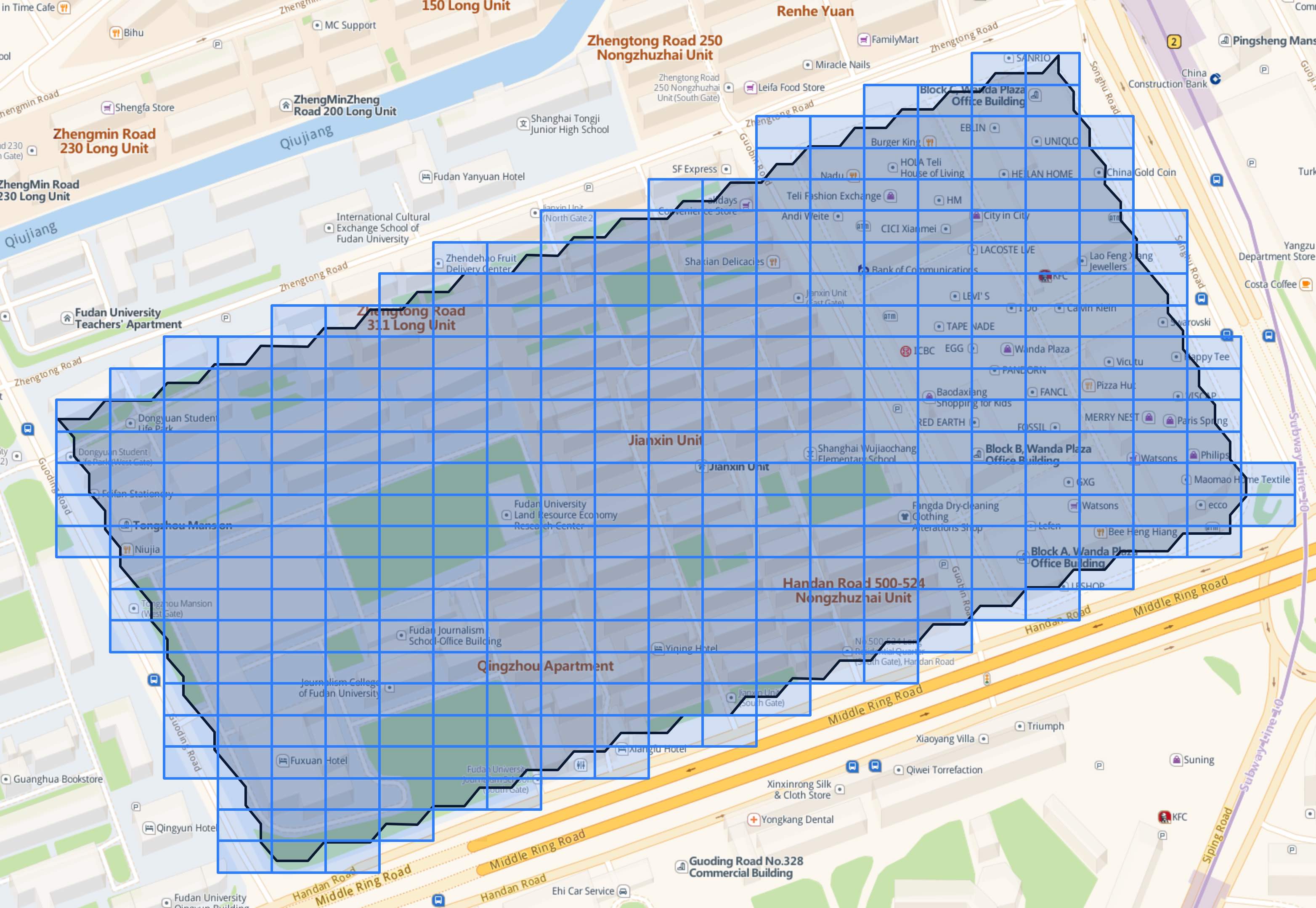}
    } 
    \caption{Examples of the serviced area and total area, the service specificity is calculated by 8-bit geohash.}
    \label{fig: geohash_details}
\end{figure}

\section{Multi-objective Node Swapping Algorithm}
\label{app:algorithm}


\SetKwFunction{FMain}{MovableBoundary}
\SetKwProg{Fn}{Function}{:}{}

\begin{algorithm}[h]
\small
\caption{Predictability-Specificity Co-optimization}\label{alg:local_search}
\LinesNumbered 
\KwIn{$D \in \mathbb{R}^{T\times N}$, $A \in \mathbb{R}^{N\times N}$, $vs,ts \in \mathbb{R}^{N\times 1}$, Max Epochs $Eps$, Geographic Constraints $\mathcal{C}$, Predictability Weight $w$}
\KwOut{Pareto optimal solutions $\mathcal{Y}$}
Initialize $bestACF \gets 0$, $bestSpecificity \gets 0$, $\mathcal{Y} \gets \emptyset$ \;
Generate feasible solutions and append them to $\mathcal{Y}$ \;
\While{$Eps > 0$}{
Sample a random number $p$ $\in [0,1]$ from uniform distribution \;
\If{$p < w$}{
    Select the solution with the best predictability $X$ from $\mathcal{Y}$ \;
}
\Else{
    Select the solution with the best specificity $X$ from $\mathcal{Y}$ \;
}
Get boundary set $\mathcal{B}\gets$ \FMain{X,A,$\mathcal{C}$} \;
\For{each $(u,v) \in \mathcal{B}$}{
    Move $u$ to the cluster of $v$ get new solution $X^{\prime}$ \;
    Calculate the gain of $X^{\prime}$ (i.e., \textit{acf} \& \textit{specificity}) \;
    \If{$\textit{acf} > \textit{bestACF}$ \textbf{or} $\textit{specificity} > \textit{bestSpecificity}$}{
        update \textit{bestACF} or \textit{bestSpecificity} \;
        append new solution $X^{\prime}$ to $\mathcal{Y}$ \;
        }
    $Eps \gets Eps - 1$ \;
    }
Remove non-Pareto solutions in $\mathcal{Y}$ \;
}
\Return $\mathcal{Y}$
\\\hrulefill \\
\Fn{\FMain{X,A,$\mathcal{C}$}}{
 Initialize $\mathcal{B} \gets \emptyset$  \;
\For{$m \text{ in } \{1,2,...,M\}$}{
    Find nodes set $U^{\prime}$ that belongs to cluster $m$ \;
    \For{each node $u \in U^{\prime}$}{
    Find neighbour set $V^{\prime}$ of node $u$ \;
    \For{each node $v \in V^{\prime}$}{
        \If{$u,v$ belong to different clusters}{
            move $u$ to the cluster of $v$ and get $X^{\prime}$ \;
            \If{constraints $\mathcal{C}$ are satisfying in $X^{\prime}$}{
                append pair of nodes $(u,v)$ to $\mathcal{B}$ \;
                }
             }
        }
    }
}
\Return $\mathcal{B}$
}
\end{algorithm}

\begin{table}[htbp]
    \caption{Dataset Statistics.}
    \resizebox{0.495\textwidth}{!}{
    \begin{tabular}{ccccccc}
        \toprule
         \textbf{Datasets} & \textbf{Designated Driver} & \textbf{Freight Transport} & \textbf{Open Taxi}  \\ 
         \midrule
         \# Records   & 8,000,000  & 7,000,000 & 9,455,822 \\
         Time Span  & 2020.10-2021.08 & 2020.10-2021.08  & 2018.10-2019.05 \\
         Open Access? & Private & Private & Public \\
         \midrule
         \multicolumn{4}{c}{\textbf{Road Data (OpenStreetMap)}} \\
         Type & 8 & 8 & N/A \\
         \# Roads & 6867 & 6867 & N/A\\
         \midrule
         \multicolumn{4}{c}{\textbf{River Data (OpenStreetMap)}}\\
         \# Rivers & 1873 & 1873 & N/A \\
        \bottomrule
    \end{tabular}}
    \label{table: dataset_statistical}
\end{table}

\section{Dataset Description} \label{data_description}

The \textbf{\textit{Designated Driver Dataset}} and \textbf{\textit{Freight Transport Dataset}} are both sampled from a world-leading online transportation company. They include the designated driver orders and the freight transport orders from Oct. 2020 to Aug. 2021 in a mega-city. The designated driver order typically takes place like this: after drinking alcoholic beverages, people are not allowed to drive and seek help from the sober designated driver to take them home. The freight transport service is similar to online ride-sharing services. That is, users send orders online, and truck owners receive orders online and provide transportation services. We sample a certain percentage of these two datasets and get a 10-month dataset with 8,000,000 and 7,000,000 records respectively. Each record contains the start time and location (longitude and latitude).


\textbf{\textit{Open Taxi Dataset}}. The taxi demand dataset is collected from Chicago's open data portal.\footnote{https://data.cityofchicago.org/Transportation/Taxi-Trips/wrvz-psew}
The dataset describes taxi trip records including the pickup time and location. Note that to protect privacy, the latitude and longitude of the trips are not recorded in the dataset; instead, the location is reported at the census tract level.
Considering that census tracts already hold good spatial semantics, we use census tracts as the atomic spatial elements for clustering (without the need to do segmentation). We obtain the polygon boundaries of the census tracts from Chicago's open data portal\footnote{https://data.cityofchicago.org/Facilities-Geographic-Boundaries/Boundaries-Census-Tracts-2010/5jrd-6zik}.

\textbf{\textit{Road and Obstacle Dataset}}.
Road and obstacle data are collected from OpenStreetMap (OSM).\footnote{We download OSM data from https://download.geofabrik.de/} 
To produce fine-grained level road segmentation, we choose the majority of vehicle roads, primarily those classified as `motorways', `primary', `secondary', and `tertiary'. We extract river records from OSM waterway data.

\section{Baselines and Experimental Setting} \label{setting_and_baselines}

\subsection{Baselines}
For a fair comparison, \textit{RegionGen} and all the baselines are tuned to generate the same number of regions ($M$). Specifically, $M$ is set to 913, 724, and 95, respectively, for the designated driver, freight transport, and open taxi datasets. The baselines are as follows.

\begin{itemize}[leftmargin=0.2cm]
\item \textbf{\textit{Grid}} and \textbf{\textit{Hexagon}}: With poor spatial semantic meaning, we split the city into several grids/hexagons of equal size. The elements without spatiotemporal data will be filtered out. 

\item \textbf{\textit{DBSCAN}}: \textit{DBSCAN} \cite{dbscan_1996} is used for clustering transportation service orders' location  points. It is a point clustering method and cannot output the shape of the generated regions directly. 
\end{itemize}
Previous research has proposed several station-based clustering methods \cite{li_traffic_2015,chen_dynamic_2016}, which aggregate nearby stations with similar spatiotemporal patterns. To adopt these methods, we take the atomic spatial elements as stations to generate regions by clustering.
\begin{itemize}[leftmargin=0.2cm]
\item \textbf{\textit{BSC}} \cite{li_traffic_2015}: The \textit{Bipartite Station Clustering} (BSC) method groups stations into clusters based on their geographical locations (first partition) and transition patterns (second partition). As our datasets include demand records, we cluster the station by the demand matrix instead of the transition matrix in the second partition. 

\item \textbf{\textit{GCSC}} \cite{chen_dynamic_2016}: The \textit{Geographically-Constrained Station Clustering} (GCSC) method groups stations into clusters, making each cluster consist of neighboring stations with similar usage patterns. 
\end{itemize}

\subsection{Experimental Setting of Region Generation} 
\label{sub:setting_region}
The road and obstacle vectors in the entire city (about $80km\times 70km$) are converted into a binary raster with $8000\times 6000$ pixels. We apply a small $5\times 5$ dilation kennel as Yuan et al. \cite{yuan2012segmentation}. In the graph generation component, atomic elements whose $\text{ACF\_daily}>0.5$ or $\text{area}>5 km^2$ are separate nodes. The geographic adjacent distance $\tau$ is 50m. The maximum area constraint is $5 km^2$. 

\subsection{Setting of Spatiotemporal Prediction} \label{sub:st_prediction}
To conduct demand prediction, we first select the spatiotemporal data in the last 10\% duration in each dataset as test data and the 10\% data before the test data as the validation test. All region generation methods are based on the data in the train set. We apply three state-of-the-art predictive models, including \textit{STMGCN} \cite{geng2019spatiotemporal}, \textit{STMeta} \cite{STMeta}, and \textit{GraphWaveNet} \cite{graphwavenet_2019}). 
To capture spatial dependences, we introduce distance and correlation graphs as Wang et al. \cite{STMeta}. The distance graphs are calculated based on the Euclidean distance while the correlation graphs are computed by the Pearson coefficient of demand series. The hyperparameters of these deep models are fine-tuned by grid search and the hidden states of the STMGCN network are 64 (the dimension of spatiotemporal representations). The degree of graph Laplacian is set to 1. We use the Adam optimizer with learning rate = 1e-4.




\section{Initialization Methods} 
\label{sec:initial_methods}

\textbf{\textit{D-Balance}}. Namely the \textit{fast solver} in Sec. \ref{cluster_scale_estimation}, we get the results by assigning the node weight with the amount of spatiotemporal data and minimize the edge-cut (to isolate the nodes
with small degrees) while tolerating 5\% unbalance.

\textbf{\textit{Greedy}}. Graph growing is a widely adopted graph partition technique \cite{gpa_graph_2007,Buluç2016}. The simplest growing method starts from a random node $v$, remaining nodes are assigned to the same cluster using a breadth-first search. The growth stops when a certain constraint transgresses. We extend the growing method by a greedy strategy to guide the node growth and optimize the objectives (e.g., predictability). First, we randomly select $M$ nodes as the initial points of $M$ clusters. Then we add unassigned nodes to the assigned clusters in turn by picking the unassigned node with the greatest gain. The following equation specifies the gain of appending node $v$:
\begin{equation}
    gain(v) = \lambda \cdot \Delta \text{ACF} + (1-\lambda) \cdot \Delta \text{Specificity}
\end{equation}
where $\Delta$ACF and $\Delta$Specificity are the change of the ACF\_daily and specificity after appending node $v$ into assigned sets. $\lambda$ (usually set to 0.7 in practice) is a trade-off parameter between choosing better predictability solutions or better specificity solutions. Therefore, the \textit{Greedy} solver converts the original problem into a single-objective optimization problem by the linear scalarization.

\textbf{\textit{Fluid}} \cite{pares2017fluid} is a community detection technique based on how fluids interact with one another and change size in their environment. We get the solutions by giving the `aggregatable' graph $G=(V, E)$ and using a propagation-based approach with predefined cluster numbers.

\end{document}